\documentclass{article}

\usepackage{microtype}
\usepackage{graphicx}
\usepackage{booktabs} 

\usepackage{hyperref}

\usepackage[accepted]{icml2026}

\usepackage{amsmath}
\usepackage{amssymb}
\usepackage{mathtools}
\usepackage{amsthm}

\usepackage{adjustbox}
\usepackage{textcomp}
\usepackage{subcaption}

\usepackage{tabularx}
\usepackage{multirow}
\usepackage{makecell}
\usepackage{bm}

\usepackage[capitalize,noabbrev]{cleveref}

\usepackage[table]{xcolor}

\theoremstyle{plain}
\newtheorem{theorem}{Theorem}[section]
\newtheorem{proposition}[theorem]{Proposition}
\newtheorem{lemma}[theorem]{Lemma}

\theoremstyle{definition}

\newtheorem{remark}[theorem]{Remark}
\theoremstyle{remark}

\usepackage[textsize=tiny]{todonotes}

\begin{document}

\twocolumn[
\icmltitle{Gauge-Equivariant Graph Networks via Self-Interference Cancellation}


\icmlsetsymbol{equal}{*}

\begin{icmlauthorlist}
\icmlauthor{Yoonhyuk Choi}{yyy}
\icmlauthor{Jiho Choi}{comp}
\icmlauthor{Jiwoo Kang}{yyy}
\end{icmlauthorlist}

\icmlaffiliation{yyy}{Department of Artificial Intelligence, Sookmyung Women's University, Seoul, South Korea}
\icmlaffiliation{comp}{Korea Advanced Institute of Science and Technology, Seoul, South Korea}

\icmlcorrespondingauthor{Jiwoo Kang}{jwkang@sookmyung.ac.kr}

\icmlkeywords{Graph neural networks, graph heterophily, self-interference cancellation, gauge-equivariant}

\vskip 0.3in
]



\printAffiliationsAndNotice{}  

\begin{abstract}
Graph Neural Networks (GNNs) excel on homophilous graphs but often fail under heterophily due to self-reinforcing and phase-inconsistent signals. We propose a \textbf{G}auge-\textbf{E}quivariant Graph Network with \textbf{S}elf-Interference \textbf{C}ancellation (GESC), which replaces additive aggregation with a projection-based interference mechanism. Unlike prior magnetic or gauge-equivariant GNNs that rely on additive message mixing, GESC explicitly models self-interference arising from redundant low-frequency components. We show that the absence of interference handling in existing gauge-based GNNs is a primary driver of oversmoothing under gauge transport. We introduce a $\mathrm{U}(1)$ phase connection followed by a rank-1 projection that suppresses self-parallel components before attention, and a sign-aware gate that regulates negatively aligned neighbors. Across diverse graph benchmarks, GESC consistently outperforms recent state-of-the-art models while offering a unified, interference-aware view of message passing. Our code is available at \href{https://github.com/ChoiYoonHyuk/GESC}{this link}.
\end{abstract}

\section{Introduction}
Graph Neural Networks (GNNs) have emerged as a powerful paradigm for learning on relational data, achieving state-of-the-art performance across social networks, recommendation, and scientific computing \citep{kipf2017semi, velivckovic2018graph, gilmer2017neural}. Despite these successes, most GNNs rely on real-valued message passing that implicitly assumes homophily, symmetric interactions, and non-negative aggregation \citep{wu2019comprehensive, bronstein2021geometric}. This inductive bias aligns well with classical benchmarks but fails to represent the directional or oscillatory nature of many real-world systems.

Many graphs exhibit interfering interactions that cannot be faithfully captured by additive aggregation. Such phenomena arise in systems governed by transport, diffusion, or wave propagation, and can be modeled through magnetic Laplacians, signed-directed operators, or high-frequency spectral components \citep{levie2018cayleynets, rusch2020coupled, marques2020signal, wang2021dissecting, lim2022sign, ko2023spectral}. However, standard message passing aggregates neighbor signals indiscriminately, amplifying redundant components and suppressing antisymmetric structure. Under this condition, repeated message-passing reinforces self-parallel information over multiple hops, which can accumulate biases (over-smoothing) \citep{li2018deeper, oono2019graph, alon2021bottleneck, black2023understanding} and degrade under heterophily \citep{zhu2020beyond, lim2021linkx}. This failure to represent destructive interference limits GNN expressivity in non-homophilous settings.

Recent studies have sought to alleviate these issues through more expressive aggregation using signed messaging \citep{huang2019sign, li2022learning, choi2025beyond}, heterophily-aware architectures \citep{jin2021universal, choiselective}, adaptive or spectral filtering \citep{chien2021adaptive, he2021bernnet, ma2021simplifying}, or decomposition-based designs \citep{wang2021tree, yan2022two}. While these models mitigate over-smoothing, they remain fundamentally scalar reweighting schemes that fail to represent the phase structure of interactions. Without explicit modeling of relative phases or destructive alignment, such approaches cannot express oscillatory or signed behaviors essential for directional and anti-homophilous graphs \citep{levie2018cayleynets, xu2019powerful, de2021gauge, brandstetter2022geometric, zheng2022heterophily, rampavsek2022graphgps}. This motivates a principled mechanism that directly models and cancels interference, rather than merely reweighting it away.

To realize this idea, our model represents nodes as complex embeddings and equips each edge with a learnable $U(1)$ phase connection with gauge-equivariant transport \citep{de2021gauge, brandstetter2022geometric}. Unlike prior magnetic Laplacian-based or gauge-equivariant GNNs \citep{zhang2021magnet, he2022msgnn}, which primarily achieve phase-consistent message passing through scalar reweighting, we introduce a projection-based self-interference cancellation that enables true gauge- and permutation-equivariant filtering within the complex message space. Our interference-aware design removes redundant components at their source and aligns neighbor messages through magnetic transport and sign-aware gating. Specifically, (i) a projection-based Self-Interference Cancellation (SIC) removes redundant components before attention; (ii) a sign-aware gate modulates neighbors based on their relative phase; and (iii) an interference-aware message passing achieves stable propagation and mitigates over-smoothing at finite depth. Our contributions are threefold:
\begin{itemize}
\item We introduce Self-Interference Cancellation (SIC), a projection-based mechanism that removes redundant self-aligned components, mitigating constructive accumulation bias and reducing over-smoothing in GNNs.
\item We propose a Sign-aware Gating with Gauge Transport mechanism that leverages complex phase alignment for smooth and stable modulation without relying on explicit direction labels.
\item We provide theoretical analysis and extensive empirical evaluation on synthetic and real-world graph benchmarks, demonstrating improved stability, heterophily robustness, and state-of-the-art performance.
\end{itemize}

\section{Related Work}

\paragraph{Graph Neural Networks and Heterophily.}
Classical GNNs are effective in homophilous regimes, but degrade on heterophilous graphs \citep{defferrard2016convolutional, kipf2017semi, velivckovic2018graph, gilmer2017neural, abu2019mixhop}. To address this, a large body of work introduces signed or adaptive propagation mechanisms \citep{derr2018sgcn, klicpera2019appnp, gasteiger2019gdc, huang2019sign, bo2021beyond, chien2021adaptive, du2022gbk, choi2025beyond}. Recent studies also propose link-level features \citep{lim2021linkx} and decomposition-based heterophily models \citep{wang2021tree, yan2022two}. Unlike these approaches, our formulation targets heterophily through geometric interference control rather than edge reweighting, explicitly cancelling redundant self-components via rank-1 projection before attention and applying sign-aware gating.

\paragraph{Complex and Spectral-Phase Representations.}
A second line of work leverages complex embeddings to encode antisymmetry \citep{trouillon2016complex, sun2019rotate}. Early spectral \citep{levie2018cayleynets} and more recent graph networks \citep{zhang2019graph, zhang2022complex} demonstrate that complex-valued filtering can naturally encode phase and orientation. Magnetic Laplacian formulations introduce complex Hermitian operators \citep{zhang2021magnet, he2022msgnn}, and recent spectral designs extend this idea to richer frequency responses \citep{rusch2020coupled}. Most prior works incorporate phase information only at the filtering level or as an auxiliary feature. In contrast, we inject phase alignment into spatial attention and guarantee gauge-invariant scoring.

\paragraph{Gauge Equivariance and Phase Transport.}
Building on these phase-based designs, gauge-equivariant networks formalize invariance to local reference frames \citep{cohen2016group, cohen2019gauge, de2021gauge}, with recent developments connecting geometric deep learning to physical principles \citep{pei2020geom, bronstein2021geometric, brandstetter2022geometric, luo2022gauge}. We reinterpret edge phases as latent directional transport to ensure scalar scores and alignments remain gauge-invariant within our gating mechanism. This treatment complements magnetic and spectral designs \citep{marques2020signal, zhang2021magnet, he2022msgnn, lim2022sign}, but differs by learning phase transport end-to-end in the spatial domain.

\paragraph{Over-smoothing and Interference.}
Deep message passing often causes representation homogenization \citep{li2018deeper, oono2019graph, alon2021bottleneck, black2023understanding}, limiting expressivity. Typical fixes include residual connections, normalization, or hop-wise feature mixing \citep{klicpera2019appnp, gasteiger2019gdc, frasca2020sign, wang2021dissecting}. Our Self-Interference Cancellation (SIC) module takes a geometric approach: it projects out the transported neighbor component parallel to the target state before attention, mitigating self-reinforcement and preserving orthogonal signals. This stands in contrast to heuristic ego-neighbor separation methods such as H$_2$GCN \citep{zhu2020beyond}, offering a principled interference control mechanism tied to phase alignment.

More details can be found in Appendix \ref{sec:comparison}.

\section{Preliminaries}\label{sec:prelim}

\paragraph{Graph structure.}
Let $\mathcal{G} = (\mathcal{V}, \mathcal{E}, X)$ denote a finite undirected attributed graph to establish notation for subsequent sections. The node feature matrix $X \in \mathbb{R}^{N \times d_{\mathrm{in}}}$ encodes $d_{\mathrm{in}}$-dimensional real-valued features for each node. The topology is represented by a symmetric adjacency matrix $A \in \{0,1\}^{N \times N}$ with $A_{ij}=A_{ji}=1$ if and only if $\{i,j\}\in\mathcal{E}$.
We denote by $D$ the diagonal degree matrix $D_{ii} = \sum_{j=1}^N A_{ij}$ and by $\mathcal{N}(i)$ the neighbor set of node $i$.
Each node $i$ has a label $y_i\in \{1,\dots,C\}$, and $Y\in\mathbb{R}^{N\times C}$ is the one-hot label matrix. Following \cite{zhu2020beyond}, the global homophily level is given by:
\begin{equation}
\label{global_homophily}
\mathcal{G}_h \coloneqq \frac{1}{|\mathcal{E}|} \sum_{\{i,j\}\in\mathcal{E}} \mathbb{I}\bigl( y_i = y_j \bigr),
\end{equation}
where lower $\mathcal{G}_h$ indicates stronger heterophily. We focus on semi-supervised node classification with a labeled subset $\mathcal{V}_L$ and unlabeled nodes $\mathcal{V}_U = \mathcal{V}\setminus\mathcal{V}_L$.

\begin{figure*}[t]
\centering
 \includegraphics[width=\textwidth]{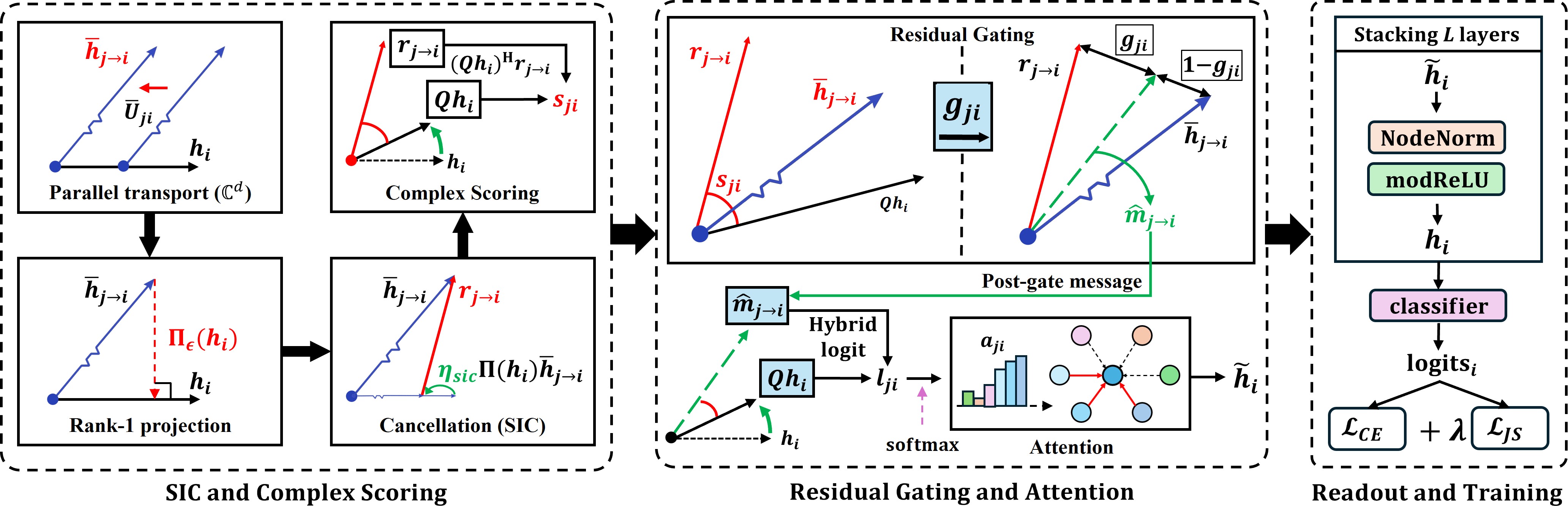}
    \caption{Overview of GESC. (\textbf{\textit{Left}}) Neighbor messages are parallel transported and self-parallel components are cancelled via rank-1 projection, yielding gauge-invariant complex scores. (\textbf{\textit{Middle}}) A sign-aware residual gate softly interpolates messages before hybrid magnitude-phase attention. (\textbf{\textit{Right}}) Aggregated features undergo NodeNorm and modReLU, followed by classification with cross-entropy and JS consistency.}
  \label{model}
\end{figure*}

\paragraph{Complex node embeddings and notation.}
We operate in a complex vector space $\mathbb{C}^d$, which naturally encodes both magnitude and phase information. Each node $i$ carries an embedding $h_i^{(t)}\in\mathbb{C}^d$ at layer $t$, where the entire node representation is defined as follows:
\begin{equation}
H^{(t)} = \begin{bmatrix} (h_1^{(t)})^\top \\ \vdots \\ (h_N^{(t)})^\top \end{bmatrix} \in \mathbb{C}^{N\times d}.
\end{equation}
The initial embedding is obtained by linearly mapping real-valued features into the complex domain using two real weight matrices for the real and imaginary parts. We denote the conjugate transpose by $(\cdot)^{\mathrm{H}}$, the elementwise complex conjugate by $\overline{(\cdot)}$, the transported quantities by $\tilde{\cdot}$ (tilde), and $\widehat{\cdot}$ (hat) for post-gating messages.

\paragraph{Complex inner products and projections.}
For the nodes $u,v\in\mathbb{C}^d$, the complex inner product is defined as
\begin{equation}
\langle u, v\rangle_{\mathbb{C}} = u^{\mathrm{H}} v 
= \sum_{k=1}^d \overline{u_k}  v_k,
\quad
\|u\|_2 = \sqrt{\langle u,u\rangle_{\mathbb{C}}}.
\end{equation}
We use a Tikhonov-regularized rank-1 projector $\Pi_\epsilon(h_i)$ onto the one-dimensional subspace spanned by $h_i$ to separate the self-parallel and orthogonal components of messages:
\begin{equation}
\label{eq_tikhonov}
\Pi_\epsilon(h_i) = \frac{h_i h_i^{\mathrm{H}}}{\|h_i\|_2^2+\epsilon},
\end{equation}
with $\epsilon>0$ for numerical stability. This operator will play a central role in the self-interference cancellation (SIC) mechanism introduced in Section \ref{sec:method}.

\paragraph{Edge transport and phase structure.}
Each undirected edge $\{i,j\}$ is assigned a unit-magnitude complex phase factor $\theta_{ji}\in\mathbb{R}$, which induces a directed pair of parallel transports with opposite phases:
\begin{equation}
\label{eq:mag_tp}
U_{ji} \coloneqq e^{ i \theta_{ji}},\quad 
U_{ij} = \overline{U_{ji}} = e^{- i \theta_{ji}},
\end{equation}
where $|U_{ji}|=|U_{ij}|=1$. We use complex weights $W, Q\in\mathbb{C}^{d\times d}$ (indices omitted here for clarity). Then, the transported source in $i$'s reference frame is given by:
\begin{equation}
\tilde{h}_{j\to i} = U_{ji} W h_j.
\end{equation}

\paragraph{Gauge-invariant formulation.}
For arbitrary phases $\{\psi_i\}$, we adopt a local $\mathrm{U}(1)$ gauge symmetry:
\begin{align}
h_i &\mapsto h'_i = e^{i\psi_i} h_i, \\ \nonumber
(\theta_{ji}, U_{ji}=e^{i\theta_{ji}}) 
&\mapsto (\theta_{ji} + \psi_i - \psi_j,\ e^{i(\psi_i - \psi_j)} U_{ji}).
\end{align}
Then, $\tilde{h}_{j\to i} \mapsto \tilde{h}'_{j\to i} = e^{i\psi_i}\tilde{h}_{j\to i}$ (gauge-equivariant). If $W,Q$ are gauge-invariant linear maps, scalar alignments like
\begin{equation}
\big(Q h_i\big)^{\mathrm{H}} \tilde{h}_{j\to i}
\end{equation}
are exactly gauge-invariant, because both factors acquire the same local phase $e^{i\psi_i}$ which cancels out in the Hermitian product. This invariance ensures that all subsequent attention logits and projection scores in Section \ref{sec:method} are computed within a locally gauge-consistent reference frame.

\section{Methodology}\label{sec:method}
\subsection{Overview}
We replace additive neighbor accumulation in GNNs with a wave-interference mechanism that (i) cancels redundant self-components before attention, (ii) softly modulates neighbor messages with a sign-aware gate to reduce harmful influence without over-suppressing negative correlations, and (iii) leverages complex-valued representations with magnitude-robust nonlinearities and optional phase-aware attention. To further stabilize sign-aware gating under noisy neighbors, we employ a magnetic transport $U_{ji}=e^{i \theta_{ji}}$ in Eq. \ref{eq:mag_tp}, which ensures gauge-invariant alignment and improves robustness in practice. All operations follow the gauge transformation convention, preserving gauge invariance of scalar scores and alignment terms.

\subsection{Proposed Method}\label{sec:gesc}
\paragraph{SIC and Complex Scoring.}
Each layer uses $M$ heads. Head $m$ applies a source transport $W^{(m)} \in \mathbb{C}^{d\times d}$ and a target filter $Q^{(m)} \in \mathbb{C}^{d\times d}$.
For each neighbor $j\in\mathcal{N}(i)$ (induced direction $j \to i$), we use the unit-modulus transport defined in Eq. \ref{eq:mag_tp}:
\begin{equation}
U_{ji} = e^{i\theta_{ji}},\quad U_{ij}=\overline{U_{ji}}.
\end{equation}
The transported source feature in $i$'s reference frame is
\begin{equation}
\tilde{h}^{(m)}_{j\to i} \coloneqq U_{ji} W^{(m)} h_j^{(t)} \in \mathbb{C}^d .
\end{equation}
To remove redundant self-parallel components before attention, we apply a Tikhonov-regularized rank-1 projector onto the target state $h_i^{(t)}$:
\begin{align}
\label{sic_impl}
\Pi_\epsilon(h_i^{(t)}) &\coloneqq \frac{h_i^{(t)}(h_i^{(t)})^{\mathrm{H}}}{\|h_i^{(t)}\|_2^2+\epsilon}, \\
r^{(m)}_{j\to i} &\coloneqq \tilde{h}^{(m)}_{j\to i} - \eta_{\mathrm{sic}}\cdot \Pi_\epsilon(h_i^{(t)}) \tilde{h}^{(m)}_{j\to i},
\end{align}
where $\eta_{\mathrm{sic}} \in [0,1]$ controls cancellation strength and $\epsilon>0$ ensures numerical stability. The complex alignment score is
\begin{equation}
s^{(m)}_{ji} \coloneqq \big(Q^{(m)} h_i^{(t)}\big)^{\mathrm{H}}  r^{(m)}_{j\to i} \in \mathbb{C}.
\end{equation}
By the $\mathrm{U}(1)$ gauge convention, $s^{(m)}_{ji}$ is a gauge-invariant scalar. For brevity, we denote the magnitude product used for normalized real alignment in the gate as
\begin{equation}
\nu^{(m)}_{ji} \coloneqq \|Q^{(m)} h_i^{(t)}\|_2 \,\|r^{(m)}_{j\to i}\|_2 + \epsilon.
\end{equation}

\paragraph{Sign-aware (phase-consistent) gating.}
We modulate the SIC-residual $r^{(m)}_{j\to i}$ by a sign-aware scalar gate $\xi^{(m)}_{ji}\in[0,1]$ computed from a gauge-invariant alignment:
\begin{align}
\rho^{(m)}_{ji} &\coloneqq 
\mathrm{Re} \left(\frac{s^{(m)}_{ji}}{\nu^{(m)}_{ji}}\right), \\ 
\xi^{(m)}_{ji} &\coloneqq \sigma \big(c_m \rho^{(m)}_{ji} + d_m\big),
\end{align}
where $c_m,d_m$ are learnable scalars and $\sigma(x)=\frac{1}{1+e^{-x}}$. The sign-aware residual is given by:
\begin{equation}
\bar r^{(m)}_{j\to i} \coloneqq \xi^{(m)}_{ji}  r^{(m)}_{j\to i}.
\end{equation}

\paragraph{Residual Gating and Alignment-based Attention.}
Each head learns a residual mixing gate $g^{(m)}_{ji}\in[0,1]$:
\begin{equation}
g^{(m)}_{ji} = \sigma \Big(a_m^\top \big[ \varphi(\|\bar r^{(m)}_{j\to i}\|_2),
                                     \varphi(\|\tilde{h}^{(m)}_{j\to i}\|_2),
                                     \varphi(|s^{(m)}_{ji}|) \big]\Big),
\end{equation}
where $\varphi(x)=\log(1+x)$. 
The post-gate message is
\begin{equation}
\label{eq:post-gate}
\widehat{m}^{(m)}_{j\to i} \coloneqq g^{(m)}_{ji}  \bar r^{(m)}_{j\to i} + \big(1-g^{(m)}_{ji}\big)  \tilde{h}^{(m)}_{j\to i}.
\end{equation}

Attention logits are computed from the complex alignment between the target and post-gate message:
\begin{equation}
\tilde{s}^{(m)}_{ji} := \big(Q^{(m)} h_i^{(t)}\big)^{\mathrm{H}}  \widehat{m}^{(m)}_{j\to i}.
\end{equation}
For the attention normalization, we use a separate magnitude product with a hybrid logit that combines magnitude and normalized real alignment:
\begin{align}
\tilde{\nu}^{(m)}_{ji} \coloneqq \|Q^{(m)} h_i^{(t)}\|_2 \,\|\widehat{m}^{(m)}_{j\to i}\|_2 + \epsilon. \\ 
\ell^{(m)}_{ji} = \gamma_m  \left[
\lambda \frac{|\tilde{s}^{(m)}_{ji}|}{\sqrt{d}} +
(1-\lambda) \mathrm{Re} \left(\frac{\tilde{s}^{(m)}_{ji}}{\tilde{\nu}^{(m)}_{ji}}\right)
\right],
\end{align}
where $\gamma_m>0$ is a learnable temperature and $\lambda\in[0,1]$ balances magnitude and phase contributions. Because $\widehat{m}^{(m)}_{j\to i}$ is obtained from gauge-equivariant $r^{(m)}_{j\to i}$ via scalar gates, $\tilde{s}^{(m)}_{ji}$ remains gauge-invariant. The attention weights and aggregated message are
\begin{align}
\label{mag_attn}
\alpha^{(m)}_{ji} &= \operatorname{softmax}_{j\in\mathcal{N}(i)} \big(\ell^{(m)}_{ji}\big),\\ 
\widetilde{h}_i^{(t+1)} &= h_i^{(t)} + \sum_{m=1}^M \sum_{j\in\mathcal{N}(i)} \alpha^{(m)}_{ji}  \widehat{m}^{(m)}_{j\to i}.
\end{align}

\paragraph{Readout and Training.}
We apply node-wise normalization and complex nonlinearity:
\begin{equation}
h_i^{(t+1)} = \mathrm{modReLU}\big(\mathrm{NodeNorm}(\widetilde{h}_i^{(t+1)})\big).
\end{equation}
Here, the $\mathrm{NodeNorm}(\cdot)$ and $\mathrm{modReLU}(\cdot)$ are given by:
\begin{align}
\mathrm{NodeNorm}(h_i) &= \frac{h_i - \mu_i}{\varsigma_i + \epsilon}, \\ 
\mathrm{modReLU}(z) &= \max(|z|+b,0)\cdot \frac{z}{|z|+\epsilon},
\end{align}
where $b$ is a (real-valued) learnable bias, $\mu_i = \tfrac{1}{d}\sum_{k=1}^d h_{ik}$, and $\varsigma_i = \sqrt{\tfrac{1}{d}\sum_{k=1}^d |h_{ik}-\mu_i|^2}$. For classification, the final complex node representations $h_i^{(L)}$ are mapped to real logits 
by concatenating their real and imaginary parts, followed by LayerNorm and a linear classifier:
\begin{align}
z_i &= \mathrm{vec}\big(\mathrm{Re}(h_i^{(L)}), \mathrm{Im}(h_i^{(L)})\big) \in \mathbb{R}^{2d}, \\ \nonumber
\mathrm{logits}_i &= W_{\mathrm{cls}}\mathrm{Dropout}(\mathrm{LayerNorm}(z_i)) \in \mathbb{R}^C.
\end{align}

We train the model with a cross-entropy loss on labeled nodes $\mathcal{V}_L$:
\begin{equation}
\mathcal{L}_{\mathrm{CE}} = - \frac{1}{|\mathcal{V}_L|} 
\sum_{i\in \mathcal{V}_L} \log p_\Theta \bigl(y_i \mid \mathrm{logits}_i\bigr),
\end{equation}
and a Jensen-Shannon consistency term between two stochastic forward passes with temperature $T>0$:
\begin{equation}
\mathcal{L}_{\mathrm{JS}} = \mathrm{JS} \left(\mathrm{softmax} \big(\mathrm{logits}^{(1)}/T\big),\ \mathrm{softmax} \big(\mathrm{logits}^{(2)}/T\big)\right).
\end{equation}
The final training objective is defined as below:
\begin{equation}
\label{eq_JS}
\mathcal{L} = \mathcal{L}_{\mathrm{CE}} + \lambda_{\mathrm{JS}} \mathcal{L}_{\mathrm{JS}}.
\end{equation}
The overall algorithm and implementation details are given in Appendix \ref{sec_alg_imp}.

\subsection{Scalability on Large Graphs} \label{sec_complexity}
For each head $m$, one GESC layer involves several computational steps. First, applying the complex linear transform $W^{(m)} \in \mathbb{C}^{d \times d}$ to all source embeddings costs $O(E d^2)$ when performed directly on edge-stacked features. Next, for each edge $(j,i)$, the SIC projection requires computing the scalar $(h_i^\ast \tilde{h}_{j\to i})$ and removing its parallel component. This involves one complex inner product and one scalar-vector multiply, resulting in $O(d)$ per edge and a total of $O(E d)$. Attention computation, including magnitude calculation and degree-wise softmax normalization, adds $O(E)$, while message formation and aggregation contribute another $O(E d)$. Summing these components and multiplying by $M$ attention heads, the overall forward complexity is $O(M E d^2)$, where the $d^2$ term arises from the complex linear transformations. Since $E$ typically scales much larger than $N$ in real-world sparse graphs and the hidden dimension $d$ remains moderate, this term dominates the runtime.

\section{Theoretical Properties}
In this section, we present theoretical justifications for the core design choices introduced in Section \ref{sec:method}. Specifically, we show that (i) Self-Interference Cancellation (SIC) attenuates self-parallel components and mitigates low-frequency dominance, (ii) sign-aware gating provides bounded and stable propagation, (iii) complex-valued transport maintains the gauge structure and invariance, and (iv) the resulting GESC layer admits explicit Lipschitz bounds, contributing to robustness against spectral collapse and over-smoothing.

\subsection{Why SIC Works: a Linear-Algebraic View}
\label{subsec:sic_spectral}
Compared to the prior works \citep{zhang2021magnet, he2022msgnn}, SIC addresses the accumulation bias, acting as a notch on low-frequency modes in self-aligned directions.

\begin{proposition}[Effect of SIC on message decomposition]\label{prop_sic_decomposition}
Let $\tilde h = \tilde h_\parallel + \tilde h_\perp$ with $\tilde h_\parallel \parallel h_i^{(t)}$ and $\tilde h_\perp \perp h_i^{(t)}$. Using the Tikhonov-regularized rank-1 operator $\Pi_\epsilon(h_i^{(t)})$ in Eq. \ref{sic_impl}, 
\begin{equation}
(I-\eta_{\mathrm{sic}}\Pi_\epsilon(h_i^{(t)}))\tilde h
= \Big(1-\eta_{\mathrm{sic}}\tfrac{\|h_i^{(t)}\|_2^2}{\|h_i^{(t)}\|_2^2+\epsilon}\Big)\tilde h_\parallel + \tilde h_\perp
\end{equation}
Thus, SIC attenuates the self-parallel component while preserving orthogonal ones. Proof is in Appendix \ref{proof_prop_sic}.
\end{proposition}

\begin{lemma}[Reduction of self-parallel energy]\label{lemma_rayleigh}
Let $\Pi_\epsilon(h_i^{(t)})$ be the Tikhonov-regularized rank-1 projector in Eq. \ref{sic_impl}, and define the self-parallel energy as follows:
\begin{equation}
\mathcal{E}_{\parallel}(x) \coloneqq \big\|\Pi_\epsilon(h_i^{(t)}) x\big\|_2^2.
\end{equation}
For any $\eta_{\mathrm{sic}}\in[0,1]$ and any $\tilde h\in\mathbb{C}^d$,
\begin{equation}
\mathcal{E}_{\parallel}\big((I-\eta_{\mathrm{sic}}\Pi_\epsilon(h_i^{(t)})) \tilde h\big)
\ \le\
\mathcal{E}_{\parallel}(\tilde h)
\end{equation}
In particular, SIC does not increase the squared magnitude of the component parallel to $h_i^{(t)}$, since $\Pi_\epsilon(h_i^{(t)})$ is a positive semidefinite rank-1 projector with operator norm at most 1. This limits the influence of self-reinforcing directions, redistributing attention mass toward more informative neighbors. Proof is in Appendix \ref{proof_lemma_rayleigh}.
\end{lemma}

\begin{remark}[Spectral notch effect]\label{rem_spectral}
Consider a linearized propagation step with normalized adjacency $\tilde A$ and self-loop weight $\alpha$:
\begin{equation}
h^{(t+1)} \approx \alpha  h^{(t)} + \tilde{A}  h^{(t)} .
\end{equation}
Under diffusion-like propagation, the dominant local direction of $h^{(t)}$ tends to align with low-frequency eigenmodes of the graph Laplacian. Applying the self-interference cancellation operator $(I-\eta_{\mathrm{sic}}\Pi_\epsilon(h_i^{(t)}))$ before attention attenuates the component of each local message parallel to $h_i^{(t)}$, reducing the energy of these low-frequency modes. Consequently, SIC acts as a node-wise notch filter that suppresses low-frequency dominance and delays spectral collapse as depth increases. A more detailed discussion is provided in Appendix \ref{proof_rem_spectral}.
\end{remark}

\subsection{Stability from Sign-aware Gating}
Sign-aware gating provides an additional mechanism for stabilizing propagation by down-weighting negatively aligned residuals before attention. Below, we prove that the resulting gated messages remain uniformly bounded in norm.

\begin{proposition}[Per-head boundedness with sign-aware gating]\label{prop_bound_xi}
Assume $U_{ji}$ is unitary, $\eta_{\mathrm{sic}}\in[0,1]$, and $\xi^{(m)}_{ji}, g^{(m)}_{ji}\in[0,1]$. Let $\{\alpha^{(m)}_{ji}\}_{j\in\mathcal N(i)}$ be the attention weights with $\alpha^{(m)}_{ji}\ge 0$ and $\sum_{j\in\mathcal N(i)} \alpha^{(m)}_{ji}=1$. Then, for any node $i$ and head $m$,
\begin{equation}
\Big\|\sum_{j\in\mathcal N(i)} \alpha^{(m)}_{ji} \widehat m^{(m)}_{j\to i}\Big\|_2
\ \le\ 
\|W^{(m)}\|_2\cdot \max_{j\in\mathcal N(i)} \|h_j^{(t)}\|_2.
\end{equation}
Proof is given in Appendix \ref{proof_prop_sign}.
\end{proposition}
This per-head bound shows that the gated aggregation cannot grow faster than a linear map with operator norm $\|W^{(m)}\|_2$, ensuring stable signal magnitudes even in deep networks. In Theorem \ref{thm_lips}, we combine this with SIC and normalization to obtain an explicit Lipschitz bound.

\subsection{Gauge Equivariance and Phase Consistency}
A central motivation for introducing complex-valued transport is to ensure the consistency under local rephasing of node embeddings. In standard message passing, scalar aggregation is sensitive to arbitrary sign or phase flips, which can destabilize attention and obscure directional relationships. By contrast, our formulation is gauge-equivariant: local phase transformations affect intermediate representations in a controlled, predictable way, and do not alter the final alignment scores or gating decisions. 

\begin{theorem}[Gauge equivariance]\label{thm_gauge}
For any choice of node-wise phases $\{\psi_i\}_{i\in\mathcal V}$, consider the local $\mathrm{U}(1)$ gauge transformation below:
\begin{equation}
h_i^{(t)} \mapsto h_i^{\prime(t)} = e^{i\psi_i} h_i^{(t)}, 
\quad
U_{ji} \mapsto U_{ji}^\prime = e^{i(\psi_i-\psi_j)} U_{ji}.
\end{equation}
For every head $m$ and edge $(j,i)$, the transported neighbor feature and the SIC residual transform covariantly as
\begin{equation}
\tilde h_{j\to i}^{\prime(m)} = e^{i\psi_i}\,\tilde h_{j\to i}^{(m)}, 
\quad
r_{j\to i}^{\prime(m)} = e^{i\psi_i}\,r_{j\to i}^{(m)}.
\end{equation}
Moreover, any scalar alignment score of the form
\begin{equation}
s^{(m)}_{ji}(v) 
= \big(Q^{(m)} h_i^{(t)}\big)^{\mathrm{H}} v^{(m)}_{j\to i},
\end{equation}
is gauge-invariant, i.e., $s^{\prime(m)}_{ji}(v') = s^{(m)}_{ji}(v)$ under transformation $v^{(m)}_{j\to i} \in \bigl\{\tilde h^{(m)}_{j\to i},\ r^{(m)}_{j\to i},\ \widehat m^{(m)}_{j\to i}\bigr\}$.
Consequently, the GESC update is gauge-equivariant in the sense that
\begin{align}
&\bigl\{h_i^{(t)}\bigr\}_{i\in\mathcal V} \mapsto \bigl\{e^{i\psi_i} h_i^{(t)}\bigr\}_{i\in\mathcal V}
\\ \nonumber
& \Longrightarrow \bigl\{h_i^{(t+1)}\bigr\}_{i\in\mathcal V} \mapsto \bigl\{e^{i\psi_i} h_i^{(t+1)}\bigr\}_{i\in\mathcal V}.
\end{align}
Proof is given in Appendix \ref{proof_thm_gauge}.
\end{theorem}
This gauge-equivariance property justifies the use of magnetic transport and complex embeddings: local phase choices at each node do not affect attention scores or gating decisions, ensuring robust and well-defined propagation across arbitrary local phase frames.

\rowcolors{2}{gray!8}{white} 
\begin{table*}[ht]
\caption{(Q1) Node classification accuracy on nine benchmark datasets with \textbf{top-3 results}. The comparison includes heterophily-aware models, advanced spectral-filtering methods, and recent state-of-the-art architectures.}
\label{tab:gesc_rq1}
\centering
\begin{adjustbox}{width=\textwidth}
\begin{tabular}{@{}lccccccccc@{}}
\Xhline{2\arrayrulewidth}
\textbf{Dataset} & \textbf{Cora} & \textbf{Citeseer} & \textbf{Pubmed} & \textbf{Actor} & \textbf{Chameleon} & \textbf{Squirrel} & \textbf{Cornell} & \textbf{Texas} & \textbf{Wisconsin} \\
\rowcolor{white}
Homophily $\mathcal{G}_h$ (Eq. \ref{global_homophily}) & 0.81 & 0.74 & 0.80 & 0.22 & 0.23 & 0.22 & 0.11 & 0.06 & 0.16 \\
\Xhline{2\arrayrulewidth}
GCN \cite{kipf2017semi}           
& 81.4$_{ \pm 0.71}$ & 67.5$_{ \pm 0.70}$ & 79.5$_{ \pm 0.47}$ & 20.3$_{ \pm 0.46}$ & 54.9$_{ \pm 0.59}$ & 31.1$_{ \pm 0.71}$ & 39.9$_{ \pm 0.79}$ & 57.0$_{ \pm 0.90}$ & 49.0$_{ \pm 0.78}$ \\
GAT \cite{velivckovic2018graph}    
& 82.6$_{ \pm 0.55}$ & 68.4$_{ \pm 0.83}$ & 79.9$_{ \pm 0.45}$ & 22.8$_{ \pm 0.41}$ & 54.4$_{ \pm 0.84}$ & 31.0$_{ \pm 0.93}$ & 42.6$_{ \pm 0.80}$ & 58.8$_{ \pm 1.01}$ & 50.2$_{ \pm 0.97}$ \\
H\textsubscript{2}GCN \cite{zhu2020beyond} 
& 80.3$_{ \pm 0.52}$ & 68.5$_{ \pm 0.76}$ & 78.8$_{ \pm 0.37}$ & 25.9$_{ \pm 1.07}$ & 53.1$_{ \pm 0.88}$ & 31.2$_{ \pm 0.68}$ & 55.0$_{ \pm 1.15}$ & 66.1$_{ \pm 1.27}$ & 62.0$_{ \pm 1.25}$ \\
SIGN \cite{frasca2020sign}      
& 82.4$_{ \pm 0.57}$ & 68.6$_{ \pm 0.70}$ & 79.7$_{ \pm 0.43}$ & 22.6$_{ \pm 0.40}$ & 54.2$_{ \pm 0.80}$ & 31.2$_{ \pm 0.94}$ & 42.4$_{ \pm 0.85}$ & 58.8$_{ \pm 0.95}$ & 50.4$_{ \pm 0.88}$ \\
GCNII \cite{chen2020simple}       
& 82.2$_{ \pm 0.64}$ & 67.8$_{ \pm 1.21}$ & 79.4$_{ \pm 0.52}$ & 26.2$_{ \pm 1.22}$ & 54.0$_{ \pm 0.77}$ & 30.8$_{ \pm 0.91}$ & 56.0$_{ \pm 1.27}$ & 69.1$_{ \pm 1.34}$ & 63.9$_{ \pm 1.29}$ \\
MagNet \cite{zhang2021magnet}   
& 83.8$_{ \pm 0.56}$ & 68.9$_{ \pm 0.67}$ & 80.1$_{ \pm 0.40}$ & 26.4$_{ \pm 0.97}$ & 56.9$_{ \pm 1.34}$ & 32.4$_{ \pm 1.15}$ & 55.1$_{ \pm 1.31}$ & 65.3$_{ \pm 1.46}$ & 61.7$_{ \pm 1.54}$ \\
GPRGNN \cite{chien2021adaptive}   
& 82.0$_{ \pm 0.59}$ & 70.1$_{ \pm 0.91}$ & 79.4$_{ \pm 0.57}$ & 25.2$_{ \pm 0.89}$ & 55.8$_{ \pm 0.81}$ & 30.6$_{ \pm 0.63}$ & 51.4$_{ \pm 1.36}$ & 60.7$_{ \pm 1.28}$ & 63.1$_{ \pm 1.21}$ \\
FAGCN \cite{bo2021beyond}  
& 82.8$_{ \pm 0.66}$ & 70.4$_{ \pm 1.20}$ & 79.8$_{ \pm 0.55}$ & 26.8$_{ \pm 1.24}$ & 54.8$_{ \pm 0.81}$ & 31.2$_{ \pm 0.87}$ & 56.8$_{ \pm 1.22}$ & 69.7$_{ \pm 1.41}$ & 64.3$_{ \pm 1.25}$ \\
ACM-GCN \cite{luan2022revisiting}  
& 82.9$_{ \pm 0.70}$ & 70.7$_{ \pm 0.81}$ & 80.0$_{ \pm 0.45}$ & 25.9$_{ \pm 1.02}$ & 56.6$_{ \pm 1.40}$ & 32.1$_{ \pm 1.05}$ & 55.1$_{ \pm 1.35}$ & 65.9$_{ \pm 1.52}$ & 62.1$_{ \pm 1.45}$ \\
GloGNN \cite{li2022finding}       
& 82.9$_{ \pm 0.45}$ & 70.9$_{ \pm 0.48}$ & \textbf{80.3$_{ \pm 0.31}$} & 27.0$_{ \pm 0.73}$ & 53.9$_{ \pm 0.70}$ & 31.0$_{ \pm 0.82}$ & 48.8$_{ \pm 1.15}$ & 62.5$_{ \pm 1.21}$ & 60.2$_{ \pm 1.12}$ \\
Auto-HeG \cite{zheng2023auto}     
& 82.4$_{ \pm 1.07}$ & 70.2$_{ \pm 1.36}$ & 80.1$_{ \pm 0.27}$ & 26.5$_{ \pm 0.99}$ & 54.3$_{ \pm 1.33}$ & 31.7$_{ \pm 1.11}$ & 53.9$_{ \pm 1.03}$ & 67.4$_{ \pm 1.65}$ & 64.0$_{ \pm 1.49}$ \\
DirGNN \cite{rossi2024edge}  
& \textbf{84.6$_{ \pm 0.61}$} & \textbf{71.5$_{ \pm 0.88}$} & 80.2$_{ \pm 0.43}$ & 27.5$_{ \pm 0.95}$ & \textbf{59.8$_{ \pm 1.45}$} & 35.2$_{ \pm 1.13}$ & \textbf{57.9$_{ \pm 1.80}$} & 68.8$_{ \pm 1.57}$ & 63.0$_{ \pm 1.33}$ \\ 
PCNet \cite{li2024pc}             
& 83.4$_{ \pm 0.77}$ & 70.8$_{ \pm 1.15}$ & 80.0$_{ \pm 0.29}$ & 26.6$_{ \pm 0.90}$ & 57.6$_{ \pm 1.65}$ & 31.8$_{ \pm 0.58}$ & 54.1$_{ \pm 1.02}$ & 62.5$_{ \pm 1.16}$ & 60.5$_{ \pm 1.13}$ \\
TFE-GNN \cite{duan2024unifying}   
& \textbf{84.1$_{ \pm 0.72}$} & \textbf{71.7$_{ \pm 1.14}$} & \textbf{80.3$_{ \pm 0.30}$} & \textbf{28.1$_{ \pm 0.81}$} & \textbf{60.2$_{ \pm 1.61}$} & \textbf{36.0$_{ \pm 0.59}$} & 53.7$_{ \pm 1.07}$ & 63.8$_{ \pm 1.11}$ & 62.5$_{ \pm 1.19}$ \\ 
CGNN \cite{pmlr-v267-zhuo25a}     
& 83.9$_{ \pm 0.70}$ & 70.5$_{ \pm 1.23}$ & 80.1$_{ \pm 0.51}$ & 26.5$_{ \pm 1.17}$ & 59.1$_{ \pm 0.78}$ & 34.4$_{ \pm 0.97}$ & \textbf{57.4$_{ \pm 1.25}$} & \textbf{70.3$_{ \pm 1.36}$} & \textbf{64.9$_{ \pm 1.22}$} \\
L2DGCN \cite{dingl2dgcn}   & 82.4$_{ \pm 0.99}$ & \textbf{71.5 $_{ \pm 0.47}$}  & 79.8 $_{ \pm 0.20}$ & \textbf{31.3$_{ \pm 0.35}$} & 53.1$_{ \pm 0.37}$  & \textbf{35.4$_{ \pm 0.52}$} & 51.5$_{ \pm 3.28}$ & \textbf{76.7$_{ \pm 2.77}$} & \textbf{65.8$_{ \pm 3.01}$} \\
\rowcolor{yellow!20}
\Xhline{2\arrayrulewidth}
\textbf{GESC (ours)} 
& \textbf{84.9$_{ \pm 0.54}$} & \textbf{72.1$_{ \pm 0.51}$} & \textbf{80.4$_{ \pm 0.10}$} & \textbf{30.4$_{ \pm 0.66}$} & \textbf{65.0$_{ \pm 1.29}$} & \textbf{37.9$_{ \pm 0.36}$} & \textbf{59.4$_{ \pm 1.93}$} & \textbf{74.5$_{ \pm 1.52}$} & \textbf{66.6$_{ \pm 2.14}$} \\
\Xhline{2\arrayrulewidth}
\end{tabular}
\end{adjustbox}
\end{table*}
\rowcolors{2}{}{}

\subsection{Stability and Non-amplification}
We now show that the interference-aware design provides explicit stability guarantees, aligning with the gating and SIC mechanisms described in Section \ref{sec:method}.

\begin{proposition}[Self-component non-amplification]\label{prop_self}
Consider the linear case (i.e., without nonlinearities or normalization) with the per-layer update below:
\begin{equation}
h_i^{(t+1)} 
= h_i^{(t)} 
+ \sum_{m=1}^M \sum_{j\in\mathcal N(i)} \alpha^{(m)}_{ji}\,\widehat m^{(m)}_{j\to i},
\end{equation}
where $\widehat m^{(m)}_{j\to i}$ is defined in Eq. \ref{eq:post-gate}. Then, for any node $i$,
\begin{align}
&\big\|\Pi_\epsilon(h_i^{(t)}) h_i^{(t+1)}\big\|_2
\ \le\ 
\big\|\Pi_\epsilon(h_i^{(t)}) h_i^{(t)}\big\|_2 \\ \nonumber
&+\sum_{m=1}^M\sum_{j\in\mathcal N(i)} \alpha^{(m)}_{ji}\,
\big\|\Pi_\epsilon(h_i^{(t)}) \tilde h^{(m)}_{j\to i}\big\|_2.
\end{align}
In particular, the SIC-processed residuals do not increase the self-parallel component. Their contribution in the direction of $h_i^{(t)}$ is always bounded above by that of the ungated transported messages, whose magnitude is controlled by the transported terms. Proof is in Appendix \ref{proof_prop_self}.
\end{proposition}

\begin{proposition}[Stability bound]\label{prop_stab}
Let $\alpha_{\max} \coloneqq \max_{m,i,j} \alpha^{(m)}_{ji}$, $\xi_{\max} \coloneqq \max_{m,i,j} \xi^{(m)}_{ji}$, and assume the in-degree is bounded by $\Delta$ (i.e., $|\mathcal N(i)|\le \Delta$ $\forall$ $i$). Then, for each head $m$ and node $i$, we can induce:
\begin{align}
\nonumber
& \Big\|
\sum_{j\in\mathcal N(i)} \alpha^{(m)}_{ji} \,\xi^{(m)}_{ji}
 \big(g^{(m)}_{ji} r^{(m)}_{j\to i} 
+ (1-g^{(m)}_{ji})\tilde h^{(m)}_{j\to i}\big)
\Big\|_2 \\ 
\ & \le\
\alpha_{\max}\,\xi_{\max}\,\Delta\, \|W^{(m)}\|_2 
\max_{j\in\mathcal N(i)}\|h_j^{(t)}\|_2.
\end{align}
This shows that soft gating controls the per-layer amplification factor at the head level, ensuring bounded signal growth. Proof is given in Appendix \ref{proof_prop_stab}.
\end{proposition}

\begin{theorem}[Lipschitz continuity and SIC-aware tightened bound]\label{thm_lips}
Assume (i) $\mathrm{NodeNorm}$ and $\mathrm{modReLU}$ are non-expansive (each is $1$-Lipschitz with respect to the $\ell_2$ norm), (ii) the magnetic transport $U_{ji}$ is unitary, and (iii) the SIC strength satisfies $0 \le \eta_{\mathrm{sic}} \le 1$ so that $\|(I-\eta_{\mathrm{sic}}\Pi_\epsilon(h_i^{(t)}))\|_2 \le 1$ for all $i,t$. Let $\alpha^{(m)}_{ji} \le \alpha_{\max}$ and suppose the in-degree is bounded by $\Delta$. Then, a single GESC layer $F$ is Lipschitz with
\begin{equation}
L_{\mathrm{GESC}}
\ \le\
1+\sum_{m=1}^M \alpha_{\max}\,\Delta\,\|W^{(m)}\|_2.
\end{equation}
Moreover, along the node-wise self-parallel direction $\mathrm{span}\{h_i^{(t)}\}$, there exist lower bounds $g_{\min},\xi_{\min},\eta_{\min}>0$ on the applied gates and SIC strength such that
\begin{equation}
L^{\parallel}_{\mathrm{GESC}} \le 1 + \sum_{m=1}^M \alpha_{\max}\,\Delta\, \big(1 - g_{\min}\,\xi_{\min}\,\eta_{\min}\big)\,\|W^{(m)}\|_2,
\end{equation}
so that the directional Lipschitz factor along self-aligned directions is strictly reduced compared to the case without SIC (i.e., $\eta_{\mathrm{sic}}=0$). In particular, SIC does not worsen the worst-case Lipschitz constant and tightens it along self-aligned directions. Further discussion of over-smoothing mitigation and complete proofs is provided in Appendix \ref{sec_oversmooth}.
\end{theorem}

\section{Experiments}
We conduct extensive experiments to evaluate our framework by addressing the following key questions:
\begin{itemize}
\item \textbf{Q1}: \textbf{Accuracy.} Does GESC improve node classification performance across both homophilous and heterophilous benchmarks?
\item \textbf{Q2}: \textbf{Ablation Study.} How do SIC, sign-aware gating, and gauge-equivariant transport individually contribute to the overall performance?
\item \textbf{Q3}: \textbf{Robustness and Sensitivity.} How sensitive is the model to variations in hyperparameters (interference cancellation rate $\eta_{\mathrm{sic}}$ and JS divergence $\lambda_{\mathrm{JS}}$) in Eq. \ref{sic_impl} and \ref{eq_JS}, and how stable is it under edge perturbations?
\item \textbf{Q4}: \textbf{Over-smoothing.} Do the proposed strategies alleviate over-smoothing and maintain stable signal propagation when stacking deeper layers?
\end{itemize}
The details of Datasets and Baselines are in Appendix \ref{det_dat_bas}.

\subsection{(Q1) Main Results}
Table \ref{tab:gesc_rq1} reports the node classification accuracy on nine benchmarks. Classical propagation models such as GCN and GAT maintain competitive accuracy on strongly homophilic graphs (e.g., Cora and Citeseer), yet their effectiveness degrades sharply when edge-label correlation weakens (e.g., Actor or Squirrel). Depth-regularized variants (GCNII and H$_2$GCN) alleviate over-smoothing but remain prone to overfitting on small-scale WebKB datasets. Methods that incorporate adaptive propagation or global re-weighting consistently outperform vanilla baselines on moderately heterophilic datasets (Cornell, Texas, Wisconsin), demonstrating the benefit of learned spectral filters and feedback aggregation. However, their advantage diminishes on WebKBs, where long-range spectral components dominate. 

Spectral architectures (DirGNN, TFE-GNN, and CGNN) achieve substantial gains in these challenging settings by leveraging edge orientation and spectral calibration to preserve cross-community dependencies. Overall, the proposed GESC exhibits the most balanced and robust performance, ranking first on seven benchmarks and within the top three on the remaining two. Its consistent accuracy across both homophilic and heterophilic graphs demonstrates the effectiveness of the proposed self-interference cancellation and sign-aware gating mechanisms, which jointly harmonize local consistency and global structural information while maintaining low variance across random splits.

\subsection{(Q2) Ablation Study}
Table \ref{tab:ablation_summary} presents the ablation results, where each major component of GESC is removed individually to assess its contribution. Removing the \textbf{Self-Interference Cancellation (w/o SIC)} significantly leads to lower accuracy across all benchmarks, verifying that eliminating self-aligned redundant signals before attention effectively mitigates over-smoothing and enhances feature diversity. The \textbf{Residual Gating (w/o RG)} mechanism also proves critical, as it adaptively balances constructive and destructive interference. Without RG, accuracy decreases by approximately $1.5\text{-}2.0\%$ on average, indicating that the residual gate prevents the amplification of harmful correlations. The absence of \textbf{Gauge-Equivariant Transport (w/o GET)} causes one of the most pronounced performance drops (up to $-3.8\%$), demonstrating that phase-consistent message transport is vital for preserving robustness under sign and orientation perturbations. Finally, replacing complex-valued representations \textbf{with real-valued ones (w/o Complex)} consistently degrades performance, confirming that the coupled magnitude-phase representation provides richer and more expressive relational features than purely real-valued aggregation.

\begin{table}[t]
\centering
\caption{(Q2) Ablation study on four benchmark datasets (Accuracy \%). Each component is incrementally removed from the full GESC model to analyze its contribution.}
\label{tab:ablation_summary}
\begin{adjustbox}{width=0.49\textwidth}
\begin{tabular}{lcccc}
\Xhline{2\arrayrulewidth}
\textbf{Variant} & \textbf{Cora} & \textbf{Citeseer} & \textbf{Chameleon} & \textbf{Texas} \\
\hline
w/o SIC        & 82.4 & 69.5 & 60.7 & 68.2 \\
w/o RG         & 83.8 & 70.0 & 62.6 & 72.4 \\
w/o GET        & 83.0 & 69.2 & 61.8 & 70.7 \\
w/o Complex    & 82.6 & 68.6 & 60.9 & 70.1 \\
\hline
\textbf{Full GESC} & \textbf{84.9} & \textbf{72.1} & \textbf{65.0} & \textbf{74.5} \\
\Xhline{2\arrayrulewidth}
\end{tabular}
\end{adjustbox}
\end{table}

\begin{figure}[ht]
     \centering
     \begin{subfigure}[b]{0.234\textwidth}
         \centering
         \includegraphics[width=\textwidth]{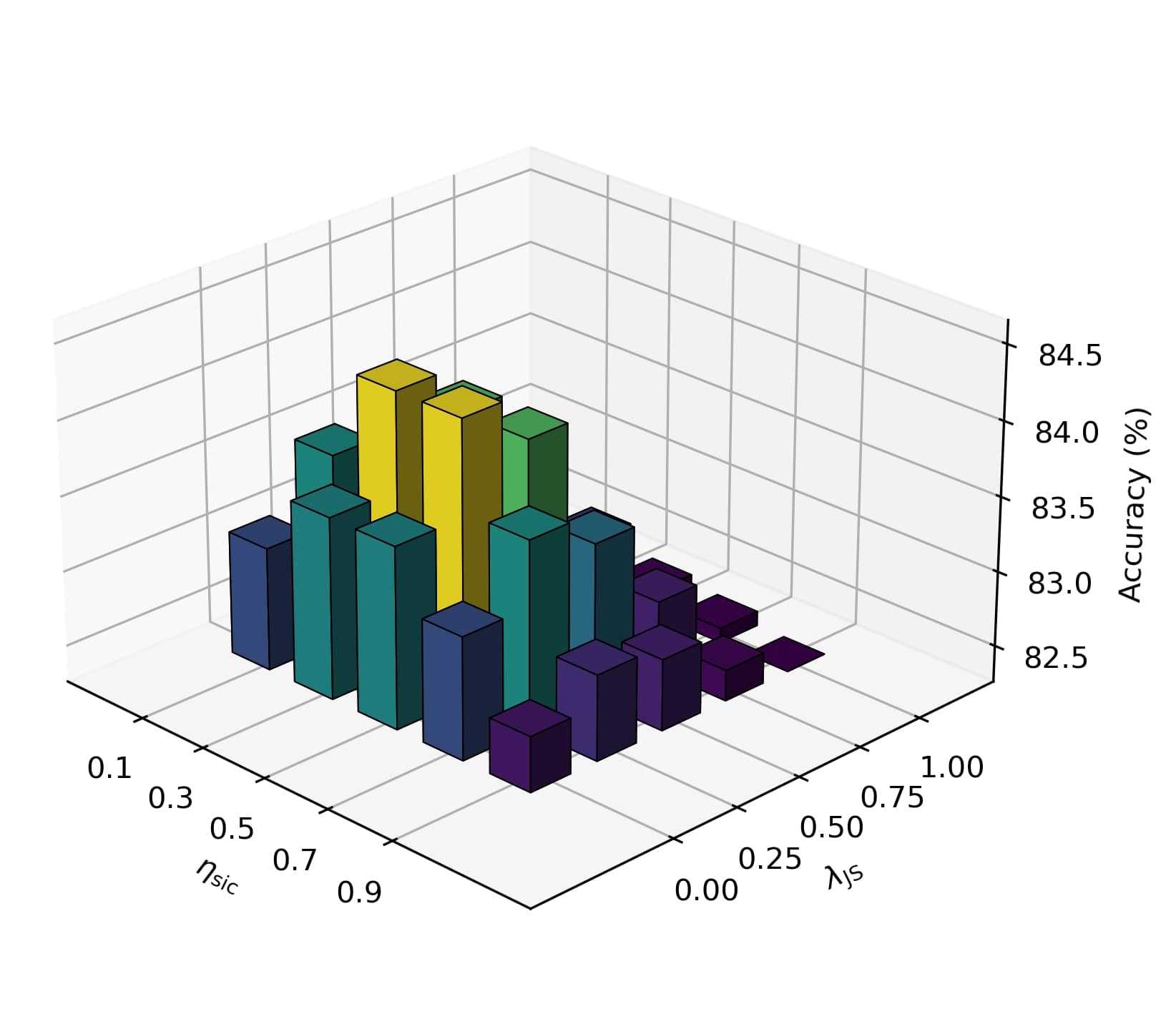}
         \caption{Cora (homophilic)}
         \label{xy_cora}
     \end{subfigure}
     \begin{subfigure}[b]{0.234\textwidth}
         \centering
         \includegraphics[width=\textwidth]{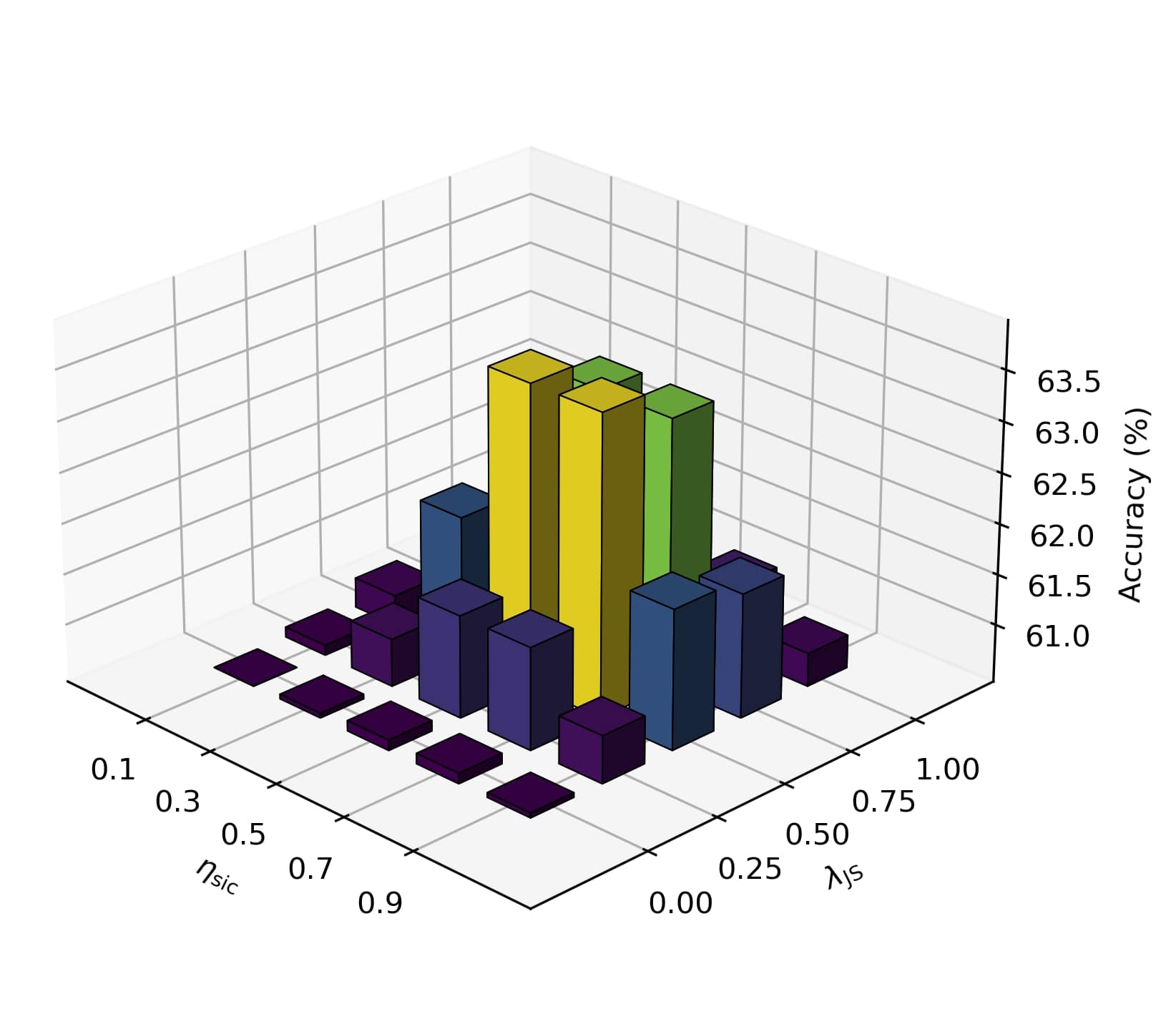}
         \caption{Chameleon (heterophilic)}
         \label{xy_cham}
     \end{subfigure}
        \caption{(Q3) Node classification accuracy to $\eta_{\mathrm{sic}}$ (Eq. \ref{sic_impl}) and $\lambda_{\mathrm{JS}}$ (Eq. \ref{eq_JS}) on two datasets. The z-axis is normalized relative to the minimum accuracy for each graph.}
        \label{xy_plot}
\end{figure}

\subsection{(Q3) Robustness and Sensitivity}
To assess the robustness of GESC, we analyze its sensitivity to two key hyperparameters: the information coupling coefficient $\eta_{\mathrm{sic}}$ and the joint-smoothing regularizer $\lambda_{\mathrm{JS}}$. Figure \ref{xy_plot} illustrates the node classification accuracy across varying $(\eta_{\mathrm{sic}}, \lambda_{\mathrm{JS}})$ settings for the Cora and Chameleon datasets. The resulting response surfaces exhibit smooth, convex-like patterns centered around the optimal region, indicating stable convergence under small hyperparameter perturbations. Even when $\eta_{\mathrm{sic}}$ or $\lambda_{\mathrm{JS}}$ deviates from its optimal value, the performance degradation remains moderate (within $1\text{--}2\%$), demonstrating that the model maintains high accuracy without requiring precise hyperparameter tuning. This stability arises from the complementary effects of self-interference cancellation and residual gating, which jointly regulate the balance between local smoothing and global consistency. Overall, these results confirm that GESC achieves robust generalization across diverse graph structures while exhibiting minimal sensitivity to hyperparameter selection.

\begin{figure}[ht]
\centering
 \includegraphics[width=.49\textwidth]{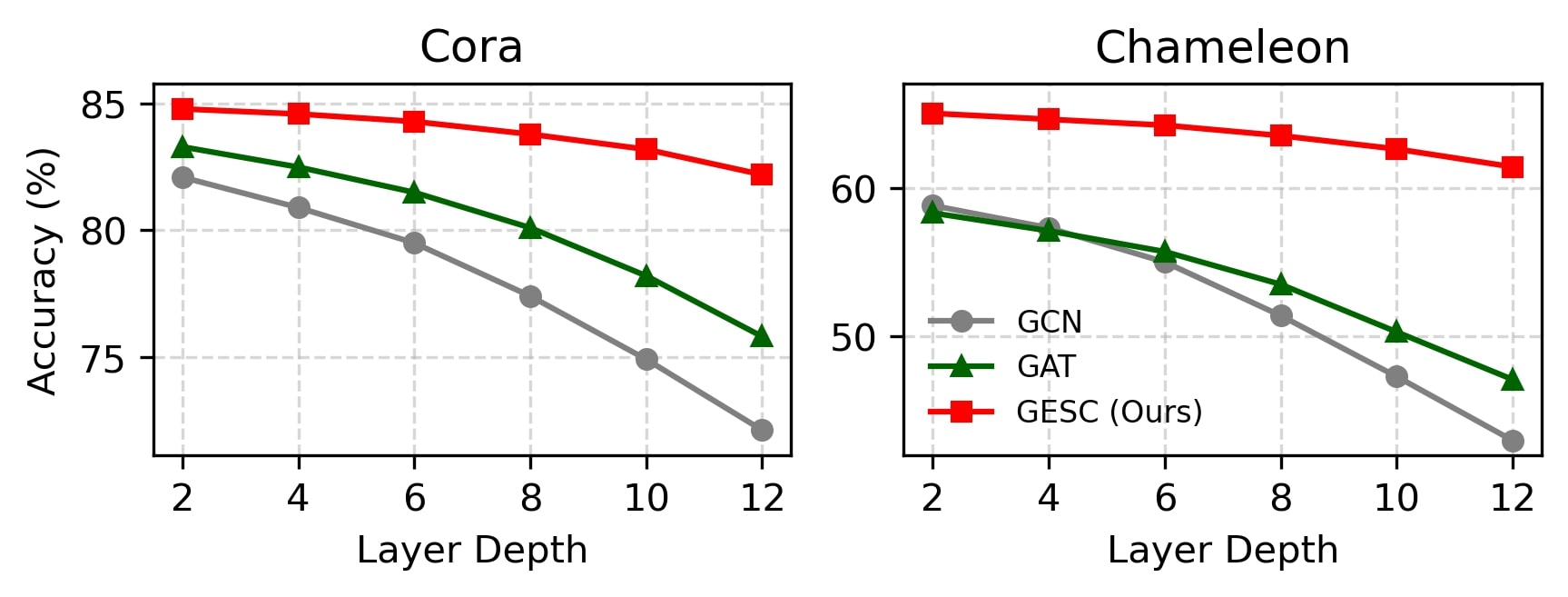}
    \caption{(Q4) Node classification accuracy versus network depth on the Cora (left) and Chameleon (right) datasets.}
  \label{fig_lifting}
\end{figure}

\subsection{(Q4) Over-smoothing}
To evaluate the effect of network depth, we analyze whether GESC mitigates the over-smoothing problem common in deep message-passing networks. As the number of layers increases, standard GCN and GAT models exhibit a sharp decline in accuracy due to feature homogenization across nodes, leading to indistinguishable embeddings. Figure \ref{fig_lifting} shows node classification accuracy on Cora and Chameleon when stacking up to 12 layers. In both datasets, GCN and GAT suffer progressive degradation with depth, confirming the accumulation of over-smoothed signals. In contrast, GESC maintains consistently higher accuracy, indicating that self-interference cancellation and residual gating effectively stabilize propagation and preserve discriminative information in deep architectures. This depth robustness arises from the interplay between local refinement and global consistency enforced by gauge-equivariant transport, preventing representation collapse as the network grows deeper.

\section{Conclusion}\label{sec:conclusion}
We presented GESC, a gauge-equivariant graph neural network that replaces additive message passing with a principled wave-interference mechanism. The self-interference cancellation suppresses redundant self-reinforcement, while sign-aware gating balances constructive and destructive interference through alignment. Theoretically, GESC guarantees gauge and permutation equivariance, mitigates over-smoothing, and maintains Lipschitz stability via spectral orthogonalization. Empirically, it achieves consistent improvements across both homophilous and heterophilous benchmarks, demonstrating robust and interpretable propagation dynamics. Beyond these results, GESC provides a geometric foundation for interference-aware graph learning, paving the way toward deeper, physically grounded models that unify topology, symmetry, and signal propagation.

\section*{Acknowledgements}
This research was supported by Sookmyung Women's University Research Grants (1-2503-2027) and by the National Research Foundation of Korea (RS-2026-25468807).

\section*{Impact Statement}
Potential benefits include more reliable node-level inference and better modeling of node interactions. GESC targets semi-supervised node classification under homophily/heterophily and is not a drop-in solution for safety-critical or privacy-sensitive tasks without additional safeguards. We do not foresee any specific negative societal consequences directly resulting from this work.

\section*{Potential Limitations}
First, GESC relies on complex-valued representations, which may introduce additional cost. Although the overall complexity remains linear in the number of edges, training can be sensitive to the nonlinearities. Second, the learned $U(1)$ transport captures phase relationships but does not directly extend to richer gauge groups. Finally, while GESC improves robustness under heterophily, its performance on extremely noisy graphs may still degrade.

\bibliography{paper}
\bibliographystyle{icml2026}

\newpage
\appendix
\onecolumn

\begin{table*}[!t]
\centering
\small
\begin{tabular}{
|>{\centering\arraybackslash}p{1.8cm}|
p{2.0cm}|
p{2.7cm}|
p{2.5cm}|
p{2.1cm}|
p{3.5cm}|
}
\hline
\textbf{Model family} &
\textbf{Method} &
\textbf{Edge operators} &
\textbf{Spectrum} &
\textbf{Interference} &
\textbf{Limitation} \\
\Xhline{1pt}

\multirow{6}{*}{\begin{tabular}[t]{c}
    Homophily \\ GNNs
  \end{tabular}} 
& GCN  & Scalar adjacency & Implicit low-pass & None & Oversmoothing \\
\cline{2-6}
& GAT  & Learned attention & Implicit & None & No heterophily modeling \\
\cline{2-6}
& GraphSAGE & Sampling-based & None & None & Limited spectral control \\
\cline{2-6}
& JKNet  & Skip connections & Implicit & None & Over-smoothing persists \\
\cline{2-6}
& DropEdge  & Edge dropout & Implicit & None & Stability issues \\
\cline{2-6}
& GCNII  & Residual & Flexible low-pass & None & Complexity increases \\
\Xhline{1pt}

\multirow{6}{*}{\begin{tabular}[t]{c}
    Heterophily \\ GNNs
  \end{tabular}} 
& H2GCN & Decoupled features & High-frequency & Partial & Heuristic interference \\
\cline{2-6}
& FAGCN & Signed filters & Adaptive spectral & Partial & No phase modeling \\
\cline{2-6}
& ACM-GCN & Adaptive mixing & Multi-hop & Partial & Sensitive to noise \\
\cline{2-6}
& MixHop & Hop mixing & Fixed spectrum & None & Limited adaptivity \\
\cline{2-6}
& GBK-GNN & Gaussian kernels & Multi-band & None & Heavy tuning \\
\cline{2-6}
& L2DGCN & Signed kernel & High-frequency & Partial & Instability on noise \\
\Xhline{1pt}

\multirow{5}{*}{\begin{tabular}[t]{c}
    Spectral \\ GNNs
  \end{tabular}}
& ChebNet & Polynomial filters & Spectral shaping & None & Fixed eigenbasis \\
\cline{2-6}
& CayleyNet & Rational filters & Flexible spectrum & None & Requires eigendecomp \\
\cline{2-6}
& BernNet & Bernstein basis & Smooth spectrum & None & High-order kernels \\
\cline{2-6}
& GPR-GNN & Polynomial filters & Flexible response & None & No interference control \\
\cline{2-6}
& JacobiConv & Jacobian filters & Spectral Jacobians & None & Linear approx only \\
\Xhline{1pt}

\multirow{3}{*}{\begin{tabular}[t]{c}
    Diffusion \\ GNNs
  \end{tabular}}
& SGC & Diffusion & Low-pass & None & Static filters \\
\cline{2-6}
& SIGN & Multi-diffusion & Fixed filters & None & No adaptivity \\
\cline{2-6}
& SignNet & Basis filters & Global structure & None & Rigid kernels \\
\Xhline{1pt}

\multirow{5}{*}{\begin{tabular}[t]{c}
    Geometric \\ GNNs
  \end{tabular}}
& NSD & Fixed sheaf maps & Sheaf Laplacian & Limited & No explicit cancellation \\
\cline{2-6}
& SheafNN & Restrictions & Sheaf spectrum & Limited & Complex maps \\
\cline{2-6}
& SheafAN & Sheaf attention & Sheaf filters & Limited & Superlinear cost \\
\cline{2-6}
& NLSD & Nonlinear sheaf & Nonlinear spectrum & Limited & Hard to train \\
\cline{2-6}
& Geom-GCN & Geometric operators & Geometric & Implicit & Weak interference control \\
\Xhline{1pt}

\multirow{2}{*}{\begin{tabular}[t]{c}
    Magnetic \\ GNNs
  \end{tabular}}
& MagNet & Magnetic Laplacian & Complex spectrum & Implicit only & Additive aggregation \\
\cline{2-6}
& MSGNN & Magnetic Laplacian & Complex phases & Implicit & No cancellation \\
\Xhline{1pt}

Gauge & GaugeCNN & Gauge-linear maps & Gauge-consistent & Implicit & No suppression of buildup \\
\cline{2-6}
\cline{2-6}
\rowcolor{yellow!20}
\cellcolor{white} GNNs & \textbf{GESC (ours)} &
\textbf{Learned U(1) + SIC} &
\textbf{Hybrid phase} &
\textbf{Explicit + sign} &
\textbf{Controls interference} \\
\hline

\end{tabular}
\caption{Comparison across model families using grouping. Each method appears in its own row, with the corresponding family shown once on the left. GESC is the only model combining gauge-invariant transport with explicit interference cancellation. The reference of each model is introduced in the following section.}
\label{tbl:comparison}
\vspace{-5pt}
\end{table*}

\section{Comparative Analysis}\label{sec:comparison}
In this section, we provide a detailed comparison between the major GNN families summarized in Table \ref{tbl:comparison}. For each family, we discuss (i) the nature of its edge operators, (ii) the resulting spectral behavior, (iii) the presence or absence of interference modeling, and (iv) the structural limitations that arise from these choices. Then, we connect these observations to the mechanisms introduced in GESC, emphasizing how Self-Interference Cancellation (SIC), sign-aware gating, and gauge-consistent transport together address the core weaknesses of prior methods.

\subsection{Classical Homophily GNNs}
Methods such as GCN \cite{kipf2017semi}, GAT \cite{velivckovic2018graph}, GraphSAGE \cite{hamilton2017inductive}, JKNet \cite{xu2018representation}, DropEdge \cite{rong2019dropedge}, and GCNII \cite{chen2020simple} rely on scalar adjacency information and adopt an additive neighborhood accumulation rule. Their propagation operators implicitly act as low-pass graph filters, repeatedly averaging node features with their neighbors. Because these models do not distinguish between self-parallel and orthogonal message components, the repeated reinforcement of the dominant low-frequency directions inevitably produces oversmoothing and loss of discriminative information. Even variants with skip connections or residual propagation preserve the same accumulation bias, as they lack any mechanism for attenuating redundant components or modeling phase-aligned interference. In contrast, GESC uses SIC to explicitly remove self-parallel redundancy before attention, thus mitigating the low-frequency collapse that characterizes this class.

\subsection{Heterophily-aware GNNs}
Models targeting heterophily: H$_2$GCN \cite{zhu2020beyond}, FAGCN \cite{bo2021beyond}, ACM-GCN \cite{luan2022revisiting}, MixHop \cite{abu2019mixhop}, GBK-GNN \cite{du2022gbk}, and L2DGCN \cite{dingl2dgcn} introduce edge operators that amplify high-frequency signals or combine multi-hop neighborhoods. While these designs improve performance on heterophilic graphs, they often rely on heuristics (e.g., signed filters or hop mixing) that provide only partial handling of interference. In particular, they lack an explicit mechanism for controlling when negatively aligned signals should be suppressed and when they carry useful information. This partial treatment causes instability in noisy settings and often requires extensive tuning. GESC directly addresses this gap: the sign-aware gate quantitatively modulates residual messages based on phase-consistent alignment, balancing suppression of harmful interference with preservation of informative high-frequency components.

\subsection{Spectral GNNs}
Spectral methods such as ChebNet \cite{defferrard2016convolutional}, CayleyNet \cite{levie2018cayleynets}, BernNet \cite{he2021bernnet}, GPR-GNN \cite{chien2021adaptive}, and JacobiConv \cite{wang2022powerful} construct polynomial or rational filters tailored to specific spectral responses. While these techniques offer precise control over the eigenvalue domain, they operate in fixed or learned spectral bases without modeling local phase interactions between neighboring nodes. As a result, they lack a notion of interference during message passing, and the filtering operation remains purely magnitude-based. Moreover, many of these models require explicit eigendecomposition or produce high-order kernels that are expensive and difficult to tune. In contrast, GESC achieves spectral adaptivity implicitly through the SIC operator and complex-valued transport, generating phase-sensitive message adjustments without the need for eigenbasis computations.

\subsection{Global Diffusion GNNs}
Global filtering approaches such as SGC \cite{wu2019simplifying}, SIGN \cite{frasca2020sign}, and SignNet \cite{lim2022sign} remove learnable propagation entirely or replace it with a small number of global diffusion channels. These frameworks are computationally efficient but inherit a strong low-pass bias from their diffusion origins and lack any form of interference control. As a consequence, they are unable to separate harmful reinforcement (e.g., self-parallel buildup) from informative global structure. GESC, by contrast, preserves learnable and local edge-sensitive behavior while maintaining robustness through SIC and gauge-consistent modulation.

\subsection{Geometric and Sheaf-based GNNs}
Geometric and sheaf-based methods: NSD \cite{bodnar2022neural}, SheafNN \cite{hansen2020sheaf}, SheafAN \cite{barbero2022sheaf}, NLSD \cite{zaghen2024sheaf} operate on richer relational structures such as sheaf Laplacians or learned geometric operators. These approaches offer enhanced expressiveness but typically suffer from limited interference modeling and often incur substantial computational overhead. Their filters are defined in sheaf or geometric spectra, but do not distinguish constructive from destructive interference across neighbors. By contrast, GESC exploits a lightweight phase-based transport and sign-aware modulation that preserves geometric consistency while remaining scalable.

\subsection{Magnetic GNNs}
MagNet \cite{zhang2021magnet}, MSGNN \cite{he2022msgnn}, and Geom-GCN \cite{pei2020geom} incorporate complex-valued transport, ensuring that messages transform consistently under local rephasing. While these models respect graph structure, they typically implement implicit-only interference handling: messages with incompatible phases are aggregated additively, with no explicit cancellation or sign-aware correction. As a result, interference may accumulate across layers and lead to unstable signal magnitudes. GESC extends this family by combining explicit SIC, sign-aware gating, and hybrid magnitude-phase attention, overcoming the buildup problem and improving robustness to noisy or misaligned neighbors.

\subsection{Why GESC is Different}
Although GaugeCNN \cite{favoni2022lattice} proposed gauge equivariant convolutional neural networks, it lacks interference cancellation and a hybrid view of messages. Comparatively, GESC is the first model to jointly incorporate: (i) Self-Interference Cancellation (SIC): a principled operator that removes redundant self-parallel components before attention, mitigating oversmoothing, (ii) sign-aware, phase-consistent gating: a gauge-invariant mechanism that suppresses harmful interference while preserving informative high-frequency contributions, and (iii) Gauge-consistent complex transport: ensuring that all alignment scores and gating decisions are invariant under local rephasing, stabilizing the propagation rule.

\section{Algorithmic and Implementation Details} \label{sec_alg_imp}

\subsection{Overall Algorithm} 

\begin{algorithm}[H]
  \caption{\textsc{GESC}: Gauge-Equivariant Self-Interference Cancellation (one training epoch)}
  \label{alg:GESC}
  \begin{algorithmic}[1]
    \REQUIRE Graph $\mathcal{G}=(\mathcal{V},\mathcal{E})$, real features $H^{(0)}_{\mathrm{real}}$, labels $Y$; hyperparameters $\{L,M,\{\gamma_m\}_{m=1}^M,\lambda,\eta_{\mathrm{sic}},\epsilon,p_{\text{edge-drop}},T,\lambda_{\mathrm{JS}}\}$
    \ENSURE Updated parameters $\Theta$

    \STATE \textbf{Complex lift:} $H^{(0)} \gets H^{(0)}_{\mathrm{real}} W_{\mathrm{re}}^{(0)} \;+\; i\, H^{(0)}_{\mathrm{real}} W_{\mathrm{im}}^{(0)}$
    \STATE \textbf{Magnetic transports:} For $(j \to i) \in \mathcal{E}$, set $U_{ji} = e^{i\theta_{ji}}$ (unit-modulus)

    \FOR{$\ell = 1$ to $L$} 
      \STATE $H^{(\ell)} \gets 0$ \COMMENT{accumulator (same shape as $H^{(\ell-1)}$)}
      \FOR{$m = 1$ to $M$} 
        \FORALL{$(j \to i)\in\mathcal{E}$}
          \STATE \textbf{Source transport and SIC projector:} $\tilde{h}^{(m)}_{j\to i} \gets U_{ji} W^{(m)} h_j^{(\ell-1)}$ $,\quad$ $\Pi_\epsilon(h_i^{(\ell-1)}) \gets \dfrac{h_i^{(\ell-1)} (h_i^{(\ell-1)})^{ H}}{\|h_i^{(\ell-1)}\|_2^2+\epsilon}$

          \STATE \textbf{Self-interference cancellation:} $r^{(m)}_{j\to i} \gets \tilde{h}^{(m)}_{j\to i} - \eta_{\mathrm{sic}}\cdot \Pi_\epsilon(h_i^{(\ell-1)}) \tilde{h}^{(m)}_{j\to i}$

          \STATE \textbf{Pre-gate complex score:} $s^{(m)}_{ji} \gets (Q^{(m)} h_i^{(\ell-1)})^{ H}\, r^{(m)}_{j\to i}$

          \STATE \textbf{Gate normalizer:} $\nu^{(m)}_{ji,\mathrm{gate}} \gets \|Q^{(m)} h_i^{(\ell-1)}\|_2\cdot \|r^{(m)}_{j\to i}\|_2 + \epsilon$

          \STATE \textbf{Phase-consistent alignment:} $\rho^{(m)}_{ji} \gets \mathrm{Re} \left(\dfrac{s^{(m)}_{ji}}{\nu^{(m)}_{ji,\mathrm{gate}}}\right)$

          \STATE \textbf{Sign-aware gate and residual:} $\xi^{(m)}_{ji} \gets \sigma \big(c_m\,\rho^{(m)}_{ji} + d_m\big)$ $,\quad$ $\bar r^{(m)}_{j\to i} \gets \xi^{(m)}_{ji}\, r^{(m)}_{j\to i}$

          \STATE \textbf{Residual gate:} $x^{(m)}_{ji} \gets \big[ \log(1+\|\bar r^{(m)}_{j\to i}\|_2),\ \log(1+\|\tilde{h}^{(m)}_{j\to i}\|_2),\ \log(1+|s^{(m)}_{ji}|) \big]$ $,\quad$ $g^{(m)}_{ji} \gets \sigma(a_m^\top x^{(m)}_{ji})$

          \STATE \textbf{Post-gate message and score:} $\widehat{m}^{(m)}_{j\to i} \gets g^{(m)}_{ji}\, \bar r^{(m)}_{j\to i} + \big(1-g^{(m)}_{ji}\big)\, \tilde{h}^{(m)}_{j\to i}$ $,\quad$ $\tilde{s}^{(m)}_{ji} \gets (Q^{(m)} h_i^{(\ell-1)})^{ H}\, \widehat{m}^{(m)}_{j\to i}$

          \STATE \textbf{Attention normalizer:} $\nu^{(m)}_{ji,\mathrm{attn}} \gets \|Q^{(m)} h_i^{(\ell-1)}\|_2\cdot \|\widehat{m}^{(m)}_{j\to i}\|_2 + \epsilon$

          \STATE \textbf{Hybrid attention logit:} 
          $\ell^{(m)}_{ji} \gets \gamma_m  \left[
          \lambda \dfrac{|\tilde{s}^{(m)}_{ji}|}{\sqrt{d}} +
          (1-\lambda)\,\mathrm{Re} \left(\dfrac{\tilde{s}^{(m)}_{ji}}{\nu^{(m)}_{ji,\mathrm{attn}}}\right)
          \right]$
        \ENDFOR

        \FORALL{$i \in \mathcal{V}$}
          \STATE \textbf{Attention:} $\alpha^{(m)}_{ji} \gets \operatorname{softmax}_{j\in\mathcal{N}(i)}\big(\ell^{(m)}_{ji}\big)$
        \ENDFOR

        \FORALL{$(j \to i)\in\mathcal{E}$}
          \STATE \textbf{Aggregate (head-$m$):} $H_i^{(\ell)} \mathrel{+}= \alpha^{(m)}_{ji}\, \widehat{m}^{(m)}_{j\to i}$
        \ENDFOR
      \ENDFOR

      \STATE \textbf{Residual update:} $\widetilde{h}_i^{(\ell)} \gets h_i^{(\ell-1)} + H_i^{(\ell)}\quad \forall i$

      \STATE \textbf{Node-wise stats (on }$\widetilde{h}_i^{(\ell)}$\textbf{):} 
      $\mu_i \gets \tfrac{1}{d}\sum_k \widetilde{h}_{ik}^{(\ell)}$, \ \ 
      $\sigma_i \gets \sqrt{\tfrac{1}{d}\sum_k |\widetilde{h}_{ik}^{(\ell)}-\mu_i|^2}$

      \STATE \textbf{NodeNorm and Complex activation:} $\mathrm{NN}(i) \gets \big(\widetilde{h}_i^{(\ell)}-\mu_i\big) / (\sigma_i+\epsilon)$ $,\quad$ $h_i^{(\ell)} \gets \mathrm{modReLU}\big(\mathrm{NN}(i)\big) \quad \forall i$
    \ENDFOR

    \STATE \textbf{Readout (real classifier):}
    \STATE \quad $z_i \gets \mathrm{vec}\big(\mathrm{Re}(h_i^{(L)}),\mathrm{Im}(h_i^{(L)})\big)$; \quad
    $\mathrm{logits}_i \gets W_{\mathrm{cls}}\ \mathrm{LayerNorm}\big(\mathrm{Dropout}(z_i)\big)$

    \STATE \textbf{Cross-entropy:} $\mathcal{L}_{\mathrm{CE}} \gets -\frac{1}{|\mathcal{V}_L|}\sum_{i\in\mathcal{V}_L}\log p_\Theta \left(y_i \mid \mathrm{logits}_i\right)$

    \STATE \textbf{JS consistency with edge dropout:}
    \STATE \quad Sample two independent edge-drop masks with prob $p_{\text{edge-drop}}$ and re-run forward to get $\mathrm{logits}^{(1)}, \mathrm{logits}^{(2)}$
    \STATE \quad $\mathcal{L}_{\mathrm{JS}} \gets \mathrm{JS} \left(\mathrm{softmax} \big(\mathrm{logits}^{(1)}/T\big),\ \mathrm{softmax} \big(\mathrm{logits}^{(2)}/T\big)\right)$

    \STATE \textbf{Total loss:} $\mathcal{L} \gets \mathcal{L}_{\mathrm{CE}} + \lambda_{\mathrm{JS}}\,\mathcal{L}_{\mathrm{JS}}$
    \STATE \textbf{Update:} $\Theta \gets \textsc{OptimizerStep}\big(\Theta, \nabla_\Theta \mathcal{L}\big)$
  \end{algorithmic}
\end{algorithm}

\subsection{Implementation Details}
All experiments are implemented in PyTorch Geometric with complex-valued extensions for message passing. We train all models using the Adam optimizer with a learning rate of $1\times 10^{-3}$ and weight decay $5\times 10^{-4}$. A single Titan XP GPU is used for all experiments. Following \cite{kipf2017semi}, we use 20 labeled nodes per class for training and randomly split the remaining nodes into validation and test sets (50\%/50\%). Early stopping is applied with patience 100 based on validation accuracy. All experiments are repeated over 10 random seeds to report the mean and standard deviation.

\paragraph{Model structure and propagation.}
Each GESC layer consists of (i) complex linear projection, (ii) gauge-equivariant transport, (iii) self-interference cancellation (SIC), and (iv) sign-aware gating with magnitude-phase attention. The transport $U_{ji}$ is represented as a phase vector and applied through elementwise complex multiplication during edge aggregation. The SIC projection is implemented as a Tikhonov-regularized rank-1 projection 
\begin{equation}
P_i^\perp = I - \eta_{\mathrm{sic}}\Pi_\epsilon(h_i),
\end{equation}
computed via a single complex inner product and outer product per edge. This operation is fused with message construction to avoid redundant memory access. Gating and attention weights are computed in parallel from real-valued scalar magnitudes and normalized with degree-wise softmax.

\paragraph{Sparse and parallel computation.}
All message passing operations are performed with sparse gather/scatter primitives. Phase transport and SIC are vectorized over edges to minimize per-edge kernel launches, and multi-head attention is parallelized across heads using fused CUDA kernels. The resulting computational complexity scales linearly with the number of edges $E$ up to the $d^2$ term from complex projections, as analyzed in Section \ref{sec_complexity}.

\section{Theoretical Proof}

\subsection{Why SIC Works: a Linear-Algebraic View}

\subsubsection{\textbf{Proof of Proposition \ref{prop_sic_decomposition}}} \label{proof_prop_sic}
Let us define $\tilde h=\tilde h_\parallel+\tilde h_\perp$ with $\tilde h_\parallel\in\mathrm{span}\{h_i^{(t)}\}$ and $\langle h_i^{(t)},\tilde h_\perp\rangle_{\mathbb C}=0$. By definition, we can easily infer that
\begin{equation}
\Pi_\epsilon(h_i^{(t)})=\frac{h_i^{(t)}h_i^{(t)H}}{\|h_i^{(t)}\|_2^2+\epsilon}
\end{equation}
is a rank-1 linear map. Since $\tilde h_\parallel = \frac{h_i^{(t)}h_i^{(t)H}}{\|h_i^{(t)}\|_2^2} \tilde h_\parallel$, we have
\begin{equation}
\Pi_\epsilon(h_i^{(t)}) \tilde h_\parallel 
= \frac{\|h_i^{(t)}\|_2^2}{\|h_i^{(t)}\|_2^2+\epsilon} \tilde h_\parallel .
\end{equation}

The orthogonality gives $h_i^{(t)H}\tilde h_\perp=0$, and thus, $\Pi_\epsilon(h_i^{(t)}) \tilde h_\perp=0$. By linearity,
\begin{equation}
(I-\eta_{\mathrm{sic}}\Pi_\epsilon(h_i^{(t)}))\tilde h
= \Big(1-\eta_{\mathrm{sic}}\tfrac{\|h_i^{(t)}\|_2^2}{\|h_i^{(t)}\|_2^2+\epsilon}\Big)\tilde h_\parallel + \tilde h_\perp,
\end{equation}
which shows that only the self-parallel component is attenuated while the orthogonal component is preserved. \qedsymbol{}

\subsubsection{\textbf{Proof of Lemma \ref{lemma_rayleigh}}} \label{proof_lemma_rayleigh}
Let $\lambda \coloneqq \frac{\|h_i^{(t)}\|_2^2}{\|h_i^{(t)}\|_2^2+\epsilon}\in(0,1)$ be the nonzero eigenvalue of $\Pi_\epsilon(h_i^{(t)})$. Since $\Pi_\epsilon(h_i^{(t)})$ is a positive semidefinite rank-1 projector, the following equality holds:
\begin{equation}
\Pi_\epsilon(h_i^{(t)})^2 = \lambda\, \Pi_\epsilon(h_i^{(t)}).
\end{equation}

Writing $\tilde h_\parallel = \Pi_\epsilon(h_i^{(t)})\tilde h$, SIC gives
\begin{equation}
\Pi_\epsilon(h_i^{(t)})(I-\eta_{\mathrm{sic}}\Pi_\epsilon(h_i^{(t)}))\tilde h
= (1-\eta_{\mathrm{sic}}\lambda)\, \tilde h_\parallel.
\end{equation}

Therefore,
\begin{equation}
\big\|\Pi_\epsilon(h_i^{(t)})(I-\eta_{\mathrm{sic}}\Pi_\epsilon(h_i^{(t)}))\tilde h\big\|_2^2
= (1-\eta_{\mathrm{sic}}\lambda)^2 \|\tilde h_\parallel\|_2^2.
\end{equation}

Since $0\le \eta_{\mathrm{sic}}\le 1$ and $0 < \lambda < 1$, we have $(1-\eta_{\mathrm{sic}}\lambda)^2 \le 1$. Injecting this inequality into the above equation leads to the following condition:
\begin{equation}
\mathcal{E}_\parallel\big((I-\eta_{\mathrm{sic}}\Pi_\epsilon(h_i^{(t)}))\tilde h\big)
\le
\mathcal{E}_\parallel(\tilde h).
\end{equation}
Thus, SIC does not increase the squared magnitude of the self-parallel component. \qedsymbol{}

\subsubsection{\textbf{Discussion for Remark \ref{rem_spectral}}} \label{proof_rem_spectral}
Let $h^{(t)}=\sum_k c_k^{(t)}u_k$ be the expansion of node features in the Laplacian eigenbasis $\{u_k\}$. For normalized adjacency $\tilde A$, we have
\begin{equation}
\tilde A u_k=\mu_k u_k,
\end{equation}
with eigenvalues $\mu_k\in[-1,1]$, where low-frequency modes correspond to larger $\mu_k$. Then, a diffusion-like propagation $h^{(t+1)} \approx \alpha  h^{(t)} + \tilde A h^{(t)}$ yields
\begin{equation}
c_k^{(t+1)} \approx (\alpha+\mu_k)c_k^{(t)},
\end{equation}
implying that low-frequency coefficients are preferentially amplified. Lastly, applying SIC introduces a local filtering term
\begin{equation}
(I - \eta_{\mathrm{sic}}\Pi_\epsilon(h_i^{(t)})),
\end{equation}
which, by Proposition \ref{prop_sic_decomposition}, suppresses components parallel to $h_i^{(t)}$. When $h^{(t)}$ has aligned with low-frequency modes, SIC locally attenuates these directions, acting as a node-wise notch filter and delaying spectral collapse. This reasoning is heuristic and reflects the frequency-selective interpretation of SIC rather than a strict global spectral theorem. \qedsymbol{}

\subsection{Stability from Sign-aware Gating}

\subsubsection{\textbf{Proof of Proposition \ref{prop_bound_xi}}} \label{proof_prop_sign}
Fix a head $m$ and a target node $i$. Recall that
\begin{equation}
\label{unitary_mp}
\tilde h^{(m)}_{j\to i} = U_{ji} W^{(m)} h_j^{(t)},
\end{equation}
with $U_{ji}$ unitary, 
\begin{equation}
r^{(m)}_{j\to i}=(I-\eta_{\mathrm{sic}}\Pi_\epsilon(h_i^{(t)})) \tilde h^{(m)}_{j\to i}, \quad
\bar r^{(m)}_{j\to i}=\xi^{(m)}_{ji}  r^{(m)}_{j\to i},
\end{equation}
and
\begin{equation}
\widehat m^{(m)}_{j\to i} = g^{(m)}_{ji} \bar r^{(m)}_{j\to i} + (1-g^{(m)}_{ji}) \tilde h^{(m)}_{j\to i},
\end{equation}
where $\xi^{(m)}_{ji}, g^{(m)}_{ji}\in[0,1]$. Given that $U_{ji}$ is unitary, the transport bound is defined as
\begin{equation}
\|\tilde h^{(m)}_{j\to i}\|_2 
= \|U_{ji} W^{(m)} h_j^{(t)}\|_2 
= \|W^{(m)} h_j^{(t)}\|_2 
\le \|W^{(m)}\|_2 \|h_j^{(t)}\|_2.
\end{equation}

Through this, we can show the non-expansiveness property of SIC using the Tikhonov-regularized rank-1 operator:
\begin{equation}
\Pi_\epsilon(h_i^{(t)})=\frac{h_i^{(t)} (h_i^{(t)})^{\mathrm{H}}}{\|h_i^{(t)}\|_2^2+\epsilon},
\end{equation}
which is Hermitian and PSD with eigenvalues $\big\{\tfrac{\|h_i^{(t)}\|_2^2}{\|h_i^{(t)}\|_2^2+\epsilon},\, 0, \dots, 0\big\}$. This property shows that $\|\Pi_\epsilon(h_i^{(t)})\|_2 \le 1$. Since the following inequality holds
\begin{equation}
\|I-\eta_{\mathrm{sic}}\Pi_\epsilon(h_i^{(t)})\|_2 \le 1,
\end{equation}
for $0\le \eta_{\mathrm{sic}}\le 1$, and thus,
\begin{equation}
\|r^{(m)}_{j\to i}\|_2 
= \|(I-\eta_{\mathrm{sic}}\Pi_\epsilon(h_i^{(t)}))\tilde h^{(m)}_{j\to i}\|_2 
\le \|\tilde h^{(m)}_{j\to i}\|_2.
\end{equation}

Now, we delve into the sign-aware scaling and residual mixing. Since $0\le \xi^{(m)}_{ji}\le 1$,
\begin{equation}
\|\bar r^{(m)}_{j\to i}\|_2 
= \|\xi^{(m)}_{ji}  r^{(m)}_{j\to i}\|_2 
\le \|r^{(m)}_{j\to i}\|_2 
\le \|\tilde h^{(m)}_{j\to i}\|_2.
\end{equation}
Moreover, $0\le g^{(m)}_{ji}\le 1$ implies that $\widehat m^{(m)}_{j\to i}$ is a convex combination of $\bar r^{(m)}_{j\to i}$ and $\tilde h^{(m)}_{j\to i}$. Consequently,
\begin{equation}
\|\widehat m^{(m)}_{j\to i}\|_2 
\le g^{(m)}_{ji}\|\bar r^{(m)}_{j\to i}\|_2 + (1-g^{(m)}_{ji})\|\tilde h^{(m)}_{j\to i}\|_2 
\le \|\tilde h^{(m)}_{j\to i}\|_2.
\end{equation}

The last term is about attention aggregation. Let $\{\alpha^{(m)}_{ji}\}_{j\in\mathcal N(i)}$ be the attention weights with $\alpha^{(m)}_{ji}\ge 0$ and $\sum_{j\in\mathcal N(i)}\alpha^{(m)}_{ji}=1$. By the triangle inequality and convexity (a weighted average is bounded by the maximum term), one can introduce:
\begin{equation}
\Big\|\sum_{j\in\mathcal N(i)} \alpha^{(m)}_{ji} \widehat m^{(m)}_{j\to i}\Big\|_2
\le \sum_{j\in\mathcal N(i)} \alpha^{(m)}_{ji} \|\widehat m^{(m)}_{j\to i}\|_2
\le \max_{j\in\mathcal N(i)} \|\widehat m^{(m)}_{j\to i}\|_2
\le \max_{j\in\mathcal N(i)} \|\tilde h^{(m)}_{j\to i}\|_2.
\end{equation}
Combining with Eq. \ref{unitary_mp} yields
\begin{equation}
\Big\|\sum_{j\in\mathcal N(i)} \alpha^{(m)}_{ji} \widehat m^{(m)}_{j\to i}\Big\|_2
\le \|W^{(m)}\|_2 \cdot \max_{j\in\mathcal N(i)} \|h_j^{(t)}\|_2,
\end{equation}
which proves the claim. \qedsymbol{}

\subsection{Gauge Equivariance and Phase Consistency}

\subsubsection{\textbf{Proof of Theorem \ref{thm_gauge}}} \label{proof_thm_gauge}
Let local $\mathrm{U}(1)$ gauge transformations act as 
\begin{equation}
h_i^{(t)} \mapsto h_i^{\prime(t)} = e^{i\psi_i} h_i^{(t)},
\quad 
U_{ji} \mapsto U_{ji}^\prime = e^{i(\psi_i-\psi_j)} U_{ji}.
\end{equation}
Assume $W^{(m)}$ and $Q^{(m)}$ are fixed (gauge-invariant) linear maps. We start with per-edge base score invariance. Consider the base transport score
\begin{equation}
\hat s_{ji}^{(m)} = \big(Q^{(m)} h_i^{(t)}\big)^{\mathrm{H}} U_{ji} W^{(m)} h_j^{(t)}.
\end{equation}
Under the gauge transform, $\hat s_{ji}^{\prime(m)}$ can be represented as
\begin{align}
\hat s_{ji}^{\prime(m)}
&= \big(Q^{(m)} h_i^{\prime(t)}\big)^{\mathrm{H}}  U_{ji}^\prime  W^{(m)} h_j^{\prime(t)} \\
&= \big(Q^{(m)} (e^{i\psi_i} h_i^{(t)})\big)^{\mathrm{H}} 
   \big(e^{i(\psi_i-\psi_j)} U_{ji}\big)  
   W^{(m)} (e^{i\psi_j} h_j^{(t)}) \\
&= e^{-i\psi_i} \big(Q^{(m)} h_i^{(t)}\big)^{\mathrm{H}}  
   e^{i(\psi_i-\psi_j)} U_{ji}  e^{i\psi_j} W^{(m)} h_j^{(t)} \\
&= \big(Q^{(m)} h_i^{(t)}\big)^{\mathrm{H}} U_{ji} W^{(m)} h_j^{(t)} 
= \hat s_{ji}^{(m)}.
\end{align}
Thus, each base per-edge score $\hat s_{ji}^{(m)}$ is gauge-invariant.

We move on to the transported message and SIC equivariance. The transported neighbor feature transforms as
\begin{equation}
\tilde h_{j\to i}^{\prime(m)}
= U_{ji}^\prime W^{(m)} h_j^{\prime(t)}
= e^{i(\psi_i-\psi_j)} U_{ji} W^{(m)} e^{i\psi_j} h_j^{(t)}
= e^{i\psi_i} \tilde h_{j\to i}^{(m)}.
\end{equation}
The $\epsilon$-regularized projector is invariant as an operator on $\mathrm{span}\{h_i^{(t)}\}$:
\begin{equation}
\Pi_\epsilon(h_i^{\prime(t)}) 
= \frac{(e^{i\psi_i} h_i^{(t)})(e^{i\psi_i} h_i^{(t)})^{\mathrm{H}}}{\|e^{i\psi_i} h_i^{(t)}\|_2^2+\epsilon}
= \frac{e^{i\psi_i} h_i^{(t)} h_i^{(t)\mathrm{H}} e^{-i\psi_i}}{\|h_i^{(t)}\|_2^2+\epsilon}
= \Pi_\epsilon(h_i^{(t)}).
\end{equation}
Thus, the SIC residual satisfies
\begin{equation}
r_{j\to i}^{\prime(m)}
= \tilde h_{j\to i}^{\prime(m)} 
  - \eta_{\mathrm{sic}} \Pi_\epsilon(h_i^{\prime(t)}) \tilde h_{j\to i}^{\prime(m)}
= e^{i\psi_i} 
  \left(\tilde h_{j\to i}^{(m)} 
  - \eta_{\mathrm{sic}} \Pi_\epsilon(h_i^{(t)}) \tilde h_{j\to i}^{(m)}\right)
= e^{i\psi_i} r_{j\to i}^{(m)}.
\end{equation}

It is also important to show the invariance of gates, logits, and the attention mechanism. Any scalar score of the form
\begin{equation}
s_{ji}^{(m)}(v) = \big(Q^{(m)} h_i^{(t)}\big)^{\mathrm{H}} v^{(m)}_{j\to i},
\end{equation}
with $v^{(m)}_{j\to i}\in\{\tilde h^{(m)}_{j\to i}, r^{(m)}_{j\to i}, \widehat m^{(m)}_{j\to i}\}$, is invariant because both arguments pick up the same phase $e^{i\psi_i}$ and the phase factors cancel in the Hermitian product. Norms $\|\cdot\|_2$ and real parts of normalized complex quantities are gauge-invariant, so all scalar functions built from them (e.g., gating logits, hybrid attention logits, and softmax-based attention weights) remain unchanged under local rephasing. Therefore, the residual gate $\xi^{(m)}_{ji}$, the mixing gate $g^{(m)}_{ji}$, and the attention weights $\alpha^{(m)}_{ji}$ are gauge-invariant. Using $r_{j\to i}^{\prime(m)}=e^{i\psi_i} r_{j\to i}^{(m)}$ and $\tilde h_{j\to i}^{\prime(m)}=e^{i\psi_i} \tilde h_{j\to i}^{(m)}$, the post-gate message transforms as
\begin{equation}
\widehat m_{j\to i}^{\prime(m)}
= g^{(m)}_{ji}\,\xi^{(m)}_{ji}\, r_{j\to i}^{\prime(m)} 
  + (1-g^{(m)}_{ji})\,\tilde h_{j\to i}^{\prime(m)}
= e^{i\psi_i} \widehat m_{j\to i}^{(m)}.
\end{equation}

Regarding layer update equivariance, the pre-activation update for node $i$ becomes
\begin{equation}
\widetilde h_i^{\prime(t+1)}
= h_i^{\prime(t)} 
  + \sum_{m=1}^M \sum_{j\in\mathcal N(i)} 
    \alpha^{(m)}_{ji} \widehat m_{j\to i}^{\prime(m)}
= e^{i\psi_i} 
  \left( h_i^{(t)} 
  + \sum_{m=1}^M \sum_{j\in\mathcal N(i)} 
    \alpha^{(m)}_{ji} \widehat m_{j\to i}^{(m)} \right)
= e^{i\psi_i} \widetilde h_i^{(t+1)}.
\end{equation}
Finally, both NodeNorm and modReLU are phase-equivariant: for any $z\in\mathbb{C}^d$ and scalar $\psi$,
\begin{equation}
\mathrm{NodeNorm}(e^{i\psi}z)=e^{i\psi}\mathrm{NodeNorm}(z),
\quad
\mathrm{modReLU}(e^{i\psi}z)=e^{i\psi}\mathrm{modReLU}(z),
\end{equation}
since both operations preserve the complex phase and act only on magnitudes. In summary, we show that the full layer update is gauge-equivariant. \qedsymbol

\subsection{Stability and Non-amplification}

\subsubsection{\textbf{Proof of Proposition \ref{prop_self}}} \label{proof_prop_self}
We consider the linearized layer update
\begin{equation}
h_i^{(t+1)} 
= h_i^{(t)} 
+ \sum_{m=1}^M \sum_{j\in\mathcal N(i)} \alpha^{(m)}_{ji}\,\widehat m_{j\to i}^{(m)},
\end{equation}
where
\begin{equation}
\widehat m_{j\to i}^{(m)} 
= g^{(m)}_{ji}\,\xi^{(m)}_{ji}\, r^{(m)}_{j\to i}
+ \big(1-g^{(m)}_{ji}\big)\,\tilde h^{(m)}_{j\to i},
\quad
r^{(m)}_{j\to i}=(I-\eta_{\mathrm{sic}}\Pi_\epsilon(h_i^{(t)})) \tilde h^{(m)}_{j\to i}.
\end{equation}
Applying the projector $\Pi_\epsilon(h_i^{(t)})$ to both sides yields
\begin{equation}
\Pi_\epsilon(h_i^{(t)}) h_i^{(t+1)}
= \Pi_\epsilon(h_i^{(t)}) h_i^{(t)}
+ \sum_{m=1}^M \sum_{j\in\mathcal N(i)} \alpha^{(m)}_{ji}\,
\Pi_\epsilon(h_i^{(t)})\widehat m_{j\to i}^{(m)} .
\end{equation}
Using the definition of $\widehat m_{j\to i}^{(m)}$,
\begin{equation}
\Pi_\epsilon(h_i^{(t)})\widehat m_{j\to i}^{(m)}
= g^{(m)}_{ji}\,\xi^{(m)}_{ji}\,\Pi_\epsilon(h_i^{(t)}) r^{(m)}_{j\to i}
+ \big(1-g^{(m)}_{ji}\big)\,\Pi_\epsilon(h_i^{(t)})\tilde h^{(m)}_{j\to i}.
\end{equation}
Taking norms and applying the triangle inequality gives
\begin{align}
\big\|\Pi_\epsilon(h_i^{(t)}) h_i^{(t+1)}\big\|_2
&\le 
\big\|\Pi_\epsilon(h_i^{(t)}) h_i^{(t)}\big\|_2
+ \sum_{m=1}^M \sum_{j\in\mathcal N(i)} \alpha^{(m)}_{ji}\,
\big\|\Pi_\epsilon(h_i^{(t)})\widehat m_{j\to i}^{(m)}\big\|_2 \\
&\le 
\big\|\Pi_\epsilon(h_i^{(t)}) h_i^{(t)}\big\|_2
+ \sum_{m=1}^M \sum_{j\in\mathcal N(i)} \alpha^{(m)}_{ji}\,
\Big( g^{(m)}_{ji}\xi^{(m)}_{ji} \big\|\Pi_\epsilon(h_i^{(t)}) r^{(m)}_{j\to i}\big\|_2
+ (1-g^{(m)}_{ji}) \big\|\Pi_\epsilon(h_i^{(t)}) \tilde h^{(m)}_{j\to i}\big\|_2 \Big).
\end{align}
By Lemma \ref{lemma_rayleigh}, SIC is non-expansive along the self-parallel direction, i.e.
\begin{equation}
\big\|\Pi_\epsilon(h_i^{(t)}) r^{(m)}_{j\to i}\big\|_2
\le \big\|\Pi_\epsilon(h_i^{(t)}) \tilde h^{(m)}_{j\to i}\big\|_2.
\end{equation}
Using $0\le g^{(m)}_{ji},\xi^{(m)}_{ji}\le1$ and combining the terms, we obtain
\begin{align}
\big\|\Pi_\epsilon(h_i^{(t)}) h_i^{(t+1)}\big\|_2
&\le 
\big\|\Pi_\epsilon(h_i^{(t)}) h_i^{(t)}\big\|_2
+ \sum_{m=1}^M \sum_{j\in\mathcal N(i)} \alpha^{(m)}_{ji}\,
\big(g^{(m)}_{ji}\xi^{(m)}_{ji} + 1-g^{(m)}_{ji}\big)\,
\big\|\Pi_\epsilon(h_i^{(t)}) \tilde h^{(m)}_{j\to i}\big\|_2 \\
&\le 
\big\|\Pi_\epsilon(h_i^{(t)}) h_i^{(t)}\big\|_2
+ \sum_{m=1}^M \sum_{j\in\mathcal N(i)} \alpha^{(m)}_{ji}\,
\big\|\Pi_\epsilon(h_i^{(t)}) \tilde h^{(m)}_{j\to i}\big\|_2,
\end{align}
since $g^{(m)}_{ji}\xi^{(m)}_{ji} + 1-g^{(m)}_{ji} \le 1$ for $g^{(m)}_{ji},\xi^{(m)}_{ji}\in[0,1]$. This yields the stated safe form: the growth of the self-parallel component is controlled by the SIC-transformed messages, and any potential increase is bounded by the contribution from the transport term. \qedsymbol{}

\subsubsection{\textbf{Proof of Proposition \ref{prop_stab}}} \label{proof_prop_stab}
For each neighbor $j$, the post-gate message is
\begin{equation}
\widehat m_{j\to i}^{(m)} 
= g^{(m)}_{ji} r_{j\to i}^{(m)} + (1-g^{(m)}_{ji})\tilde h_{j\to i}^{(m)}.
\end{equation}
Thus,
\begin{equation}
\|\widehat m_{j\to i}^{(m)}\|_2 
\le g^{(m)}_{ji}\|r_{j\to i}^{(m)}\|_2 
+ (1-g^{(m)}_{ji})\|\tilde h_{j\to i}^{(m)}\|_2.
\end{equation}
Since $U_{ji}$ is unitary and $r^{(m)}_{j\to i}$ and $\tilde h^{(m)}_{j\to i}$ are obtained by applying $(I-\eta_{\mathrm{sic}}\Pi_\epsilon(h_i^{(t)}))$ and $U_{ji}W^{(m)}$ to $h_j^{(t)}$, we have
\begin{equation}
\|r_{j\to i}^{(m)}\|_2 \le \|\tilde h_{j\to i}^{(m)}\|_2 
= \|U_{ji} W^{(m)} h_j^{(t)}\|_2
= \|W^{(m)} h_j^{(t)}\|_2
\le \|W^{(m)}\|_2 \|h_j^{(t)}\|_2.
\end{equation}
Therefore,
\begin{equation}
\|\widehat m_{j\to i}^{(m)}\|_2 
\le \|W^{(m)}\|_2 \|h_j^{(t)}\|_2.
\end{equation}

Now consider the gated aggregation with sign-aware gate:
\begin{equation}
S_i^{(m)} \coloneqq 
\sum_{j\in\mathcal N(i)} \alpha^{(m)}_{ji}\,\xi^{(m)}_{ji}
\big(g^{(m)}_{ji} r^{(m)}_{j\to i} 
+ (1-g^{(m)}_{ji})\tilde h^{(m)}_{j\to i}\big)
= \sum_{j\in\mathcal N(i)} \alpha^{(m)}_{ji}\,\xi^{(m)}_{ji}\,\widehat m^{(m)}_{j\to i}.
\end{equation}
By the triangle inequality and the bounds $\alpha^{(m)}_{ji}\le \alpha_{\max}$, $\xi^{(m)}_{ji}\le\xi_{\max}$,
\begin{align}
\|S_i^{(m)}\|_2
&\le \sum_{j\in\mathcal N(i)} \alpha^{(m)}_{ji}\,\xi^{(m)}_{ji}\,
\|\widehat m^{(m)}_{j\to i}\|_2 \\
&\le \alpha_{\max}\,\xi_{\max} \sum_{j\in\mathcal N(i)} 
\|\widehat m^{(m)}_{j\to i}\|_2 \\
&\le \alpha_{\max}\,\xi_{\max}\, |\mathcal N(i)|\, 
\max_{j\in\mathcal N(i)} \|\widehat m^{(m)}_{j\to i}\|_2.
\end{align}
Using $|\mathcal N(i)|\le \Delta$ and the bound on $\|\widehat m^{(m)}_{j\to i}\|_2$ established above,
\begin{equation}
\|S_i^{(m)}\|_2
\le \alpha_{\max}\,\xi_{\max}\,\Delta\, \|W^{(m)}\|_2 
\max_{j\in\mathcal N(i)}\|h_j^{(t)}\|_2,
\end{equation}
which is the claimed stability bound. \qedsymbol{}

\subsubsection{\textbf{Proof of Theorem \ref{thm_lips}}} \label{proof_thm_lips}
We analyze the Lipschitz constant of one GESC layer $F$ under the assumptions of unitary magnetic transport and non-expansive normalization/activation. Let us begin by bounding the single message. For any edge $(j,i)$ and head $m$, the transported message is
\begin{equation}
\tilde h_{j\to i}^{(m)} = U_{ji} W^{(m)} h_j^{(t)},
\end{equation}
where $U_{ji}$ is unitary. Thus,
\begin{equation}
\|\tilde h_{j\to i}^{(m)}\|_2 
= \|U_{ji} W^{(m)} h_j^{(t)}\|_2
= \|W^{(m)} h_j^{(t)}\|_2
\le \|W^{(m)}\|_2 \|h_j^{(t)}\|_2.
\end{equation}
After applying the self-interference cancellation (SIC),
\begin{equation}
r_{j\to i}^{(m)} = (I-\eta_{\mathrm{sic}}\Pi_\epsilon(h_i^{(t)})) \tilde h_{j\to i}^{(m)} .
\end{equation}
Since $0\le \eta_{\mathrm{sic}}\le1$ and $\Pi_\epsilon$ is a scaled rank-1 projector with operator norm at most $1$, we have $\|(I-\eta_{\mathrm{sic}}\Pi_\epsilon(h_i^{(t)}))\|_2\le1$. Therefore, $\|\tilde h_{j\to i}^{(m)}\|_2$ is bounded by
\begin{equation}
\|r_{j\to i}^{(m)}\|_2 \le \|\tilde h_{j\to i}^{(m)}\|_2 
\le \|W^{(m)}\|_2 \|h_j^{(t)}\|_2.
\label{lip_r_revised}
\end{equation}

To analyze the effect of gated message and aggregation, let the sign-aware gate be $\xi^{(m)}_{ji}\in[0,1]$ and the residual mixing gate be $g^{(m)}_{ji}\in[0,1]$. 
The post-gate message is
\begin{equation}
\widehat m^{(m)}_{j\to i} 
= g^{(m)}_{ji}\big(\xi^{(m)}_{ji} r_{j\to i}^{(m)}\big)
+ (1-g^{(m)}_{ji}) \tilde h_{j\to i}^{(m)} .
\end{equation}
By the triangle inequality and Eq. \ref{lip_r_revised},
\begin{equation}
\|\widehat m^{(m)}_{j\to i}\|_2 
\le g^{(m)}_{ji}\xi^{(m)}_{ji}\|r_{j\to i}^{(m)}\|_2 
+ (1-g^{(m)}_{ji})\|\tilde h_{j\to i}^{(m)}\|_2
\le \|W^{(m)}\|_2 \|h_j^{(t)}\|_2 .
\label{lip_m_revised}
\end{equation}

The per-head aggregated update before normalization and activation is
\begin{equation}
\widetilde h_i^{(t+1)} 
= h_i^{(t)} + \sum_{m=1}^M\sum_{j\in\mathcal N(i)} \alpha^{(m)}_{ji} \widehat m^{(m)}_{j\to i},
\end{equation}
where $\alpha^{(m)}_{ji}\ge0$ and $\sum_{j\in\mathcal N(i)} \alpha^{(m)}_{ji}\le \alpha_{\max}\Delta$ by assumption. Applying Eq. \ref{lip_m_revised} yields
\begin{equation}
\|\widetilde h_i^{(t+1)} - h_i^{(t)}\|_2
\le \sum_{m=1}^M \sum_{j\in\mathcal N(i)} \alpha^{(m)}_{ji} \|\widehat m^{(m)}_{j\to i}\|_2
\le \sum_{m=1}^M \alpha_{\max}\Delta \|W^{(m)}\|_2 
\max_{j\in\mathcal N(i)}\|h_j^{(t)}\|_2 .
\label{lip_pre_revised}
\end{equation}
\textbf{Remark.} If $\operatorname{softmax}$ attention is used per head, then $\sum_{j}\alpha^{(m)}_{ji}=1$, and $\alpha_{\max}\Delta$ can be replaced by $1$ for a tighter bound.

To show the effect of SIC and gates on the self-parallel direction, we decompose $\tilde h^{(m)}_{j\to i} = \tilde h_\parallel^{(m)} + \tilde h_\perp^{(m)}$ along $\mathrm{span}\{h_i^{(t)}\}$ and its orthogonal complement. By Proposition \ref{prop_sic_decomposition},
\begin{equation}
(I-\eta_{\mathrm{sic}}\Pi_\epsilon(h_i^{(t)})) \tilde h^{(m)}_{j\to i}
= (1-\eta_{\mathrm{sic}}\lambda_{i}^{(t)})\tilde h_\parallel^{(m)} + \tilde h_\perp^{(m)},
\end{equation}
where $\lambda_{i}^{(t)} = \frac{\|h_i^{(t)}\|_2^2}{\|h_i^{(t)}\|_2^2+\epsilon}$ is the nonzero eigenvalue of $\Pi_\epsilon(h_i^{(t)})$. Applying the projector $\Pi_\epsilon(h_i^{(t)})$ and the gates, the parallel component of the post-gate message is scaled by
\begin{equation}
g^{(m)}_{ji}\xi^{(m)}_{ji}(1-\eta_{\mathrm{sic}}) + (1-g^{(m)}_{ji})
= 1 - g^{(m)}_{ji}\xi^{(m)}_{ji}\eta_{\mathrm{sic}}.
\end{equation}
Let $g_{\min},\xi_{\min},\eta_{\min}$ be the minimal gate and SIC strengths across all heads and edges. With the slight abuse of notation, we omit the strength of the SIC here $\eta_{\min} := \eta_{\min}\lambda_{\min}$. Then,
\begin{equation}
\big\|\Pi_\epsilon(h_i^{(t)})\,\widehat m^{(m)}_{j\to i}\big\|_2
\le (1 - g_{\min}\xi_{\min}\eta_{\min})\,
\big\|\Pi_\epsilon(h_i^{(t)})\,\tilde h^{(m)}_{j\to i}\big\|_2.
\end{equation}
Aggregating across neighbors and heads yields a directional Lipschitz bound along $\mathrm{span}\{h_i^{(t)}\}$,
\begin{equation}
L_{\mathrm{GESC}}^{\parallel}
\le
1+\sum_{m=1}^M \alpha_{\max}\Delta
\big(1-g_{\min}\xi_{\min}\eta_{\min}\big)\|W^{(m)}\|_2 .
\end{equation}

The following inequality shows that the normalization $\mathrm{NodeNorm}$ and activation $\mathrm{modReLU}$ are assumed 1-Lipschitz (non-expansive)
\begin{equation}
\|\mathrm{NodeNorm}(x_1)-\mathrm{NodeNorm}(x_2)\|_2
\le \|x_1-x_2\|_2, 
\quad
\|\mathrm{modReLU}(z_1)-\mathrm{modReLU}(z_2)\|_2
\le \|z_1-z_2\|_2.
\end{equation}
Thus, they do not increase the Lipschitz constant.

Lastly, combining Eq. \ref{lip_pre_revised} with the identity path $h_i^{(t)} \mapsto h_i^{(t)}$ gives the global Lipschitz bound
\begin{equation}
L_{\mathrm{GESC}}\ \le\ 
1+\sum_{m=1}^M \alpha_{\max} \Delta \|W^{(m)}\|_2 ,
\end{equation}
and the directional bound along the self-parallel subspace,
\begin{equation}
L_{\mathrm{GESC}}^{\parallel}
\ \le\
1+\sum_{m=1}^M \alpha_{\max}\Delta
\big(1-g_{\min}\xi_{\min}\eta_{\min}\big)\|W^{(m)}\|_2 .
\end{equation}
Therefore, GESC layers are globally Lipschitz-bounded, and the directional Lipschitz factor along self-aligned directions is strictly reduced compared to the case without SIC whenever $g_{\min}\xi_{\min}\eta_{\min}>0$, which improves gradient stability and mitigates over-smoothing. \qedsymbol{}

\subsection{Over-smoothing mitigation} \label{sec_oversmooth}
Consider a GESC layer composed of Self-Interference Cancellation (SIC), sign-aware gating, and gauge-equivariant transport. Under the conditions of Propositions \ref{prop_self} and \ref{prop_stab}, and Theorem \ref{thm_lips}, the layer satisfies the following properties.

By Proposition \ref{prop_self}, the component of each node state aligned with its previous representation obeys the bound
\begin{equation}
\big\|\Pi_\epsilon(h_i^{(t)}) h_i^{(t+1)}\big\|_2
\ \le\ 
\big\|\Pi_\epsilon(h_i^{(t)}) h_i^{(t)}\big\|_2
+
\sum_{m=1}^M\sum_{j\in\mathcal N(i)} \alpha^{(m)}_{ji}\,
\big\|\Pi_\epsilon(h_i^{(t)}) \tilde h^{(m)}_{j\to i}\big\|_2.
\end{equation}
In particular, since SIC contracts the self-parallel part of each transported message, the residual branch does not increase the self-parallel component more than what would be induced by aggregating the ungated transported messages. This prevents uncontrolled reinforcement of self-loops that typically lead to feature homogenization.

From Proposition \ref{prop_stab}, the magnitude of the gated aggregation for each head $m$ and node $i$ is bounded by
\begin{equation}
\Big\|
\sum_{j\in\mathcal N(i)} 
\alpha^{(m)}_{ji}\,\xi^{(m)}_{ji}
\big(g^{(m)}_{ji}r^{(m)}_{j\to i}
+ (1-g^{(m)}_{ji})\tilde h^{(m)}_{j\to i}\big)
\Big\|_2
\ \le\
\alpha_{\max}\,\xi_{\max}\,\Delta\,
\|W^{(m)}\|_2
\max_{j\in\mathcal N(i)}\|h_j^{(t)}\|_2.
\end{equation}
Thus, each head contributes a uniformly bounded update whose growth is controlled by the operator norm of $W^{(m)}$ and the maximal in-degree, ruling out layer-wise blow-up of signal magnitudes.

By Theorem \ref{thm_lips}, GESC layers are globally Lipschitz-continuous:
\begin{equation}
L_{\mathrm{GESC}}
\ \le\
1+\sum_{m=1}^M \alpha_{\max}\,\Delta\, \|W^{(m)}\|_2,
\end{equation}
and along the self-parallel subspace,
\begin{equation}
L_{\mathrm{GESC}}^{\parallel}
\ \le\
1+\sum_{m=1}^M
\alpha_{\max}\,\Delta\,
\big(1-g_{\min}\,\xi_{\min}\,\eta_{\min}\big)
\|W^{(m)}\|_2,
\end{equation}
for lower bounds $g_{\min},\xi_{\min},\eta_{\min}>0$ on the effective gates and SIC strength. The factor $(1-g_{\min}\xi_{\min}\eta_{\min})<1$ shows that, compared with a vanilla linear propagation without SIC or gating, the effective amplification along self-aligned directions is strictly reduced. In this sense, SIC and sign-aware gating tighten the Lipschitz bound precisely where over-smoothing is most likely to occur.

\textbf{Spectral implication (heuristic).}
Let $H^{(t)}$ denote the node embedding matrix and $\mathcal{L}$ the graph Laplacian with eigen-decomposition $\mathcal{L}=U\Lambda U^\top$. Under diffusion-like propagation, low-frequency modes dominate since $\tilde A u_k \approx (1-\lambda_k)u_k$ with small $\lambda_k$. The SIC and sign-aware gating mechanisms preferentially attenuate components aligned with locally dominant directions $h_i^{(t)}$, which, in diffusion regimes, tend to correlate with low-frequency eigenmodes. Heuristically, one can interpret a GESC layer as inducing a frequency-wise damping of the form
\begin{equation}
\|U^\top H^{(t+1)}\|_F^2
\ \approx\
\sum_{k}
\big(1 - c_{\mathrm{sic}}(k)\big)^2 
\big\|U^\top H^{(t)}\big\|_{F,k}^2,
\end{equation}
where $c_{\mathrm{sic}}(k)$ grows with the effective SIC strength $g_{\min}\xi_{\min}\eta_{\min}$ and the corresponding Laplacian eigenvalue $\lambda_k$. This suggests that low-frequency energy is selectively damped while higher-frequency (structurally discriminative) modes are better preserved.

\textbf{Summary.} Together, (i)-(iv) indicate that GESC mitigates spectral collapse and over-smoothing: the self-parallel component is controlled and can be made non-increasing, the per-layer update norm remains uniformly bounded, and low-frequency spectral modes are selectively suppressed by SIC and sign-aware gating, leading to stable and diversity-preserving propagation even in deep stacks.

\begin{table*}[t]
\caption{Statistics of the nine graph datasets.}
\label{dataset}
\centering
\begin{adjustbox}{}
\begin{tabular}{@{}ccccccccccc}
&     &        &         &  & & & \\ 
\Xhline{2\arrayrulewidth}
        & \textbf{Datasets}         & \textbf{Cora}  & \textbf{Citeseer} & \textbf{Pubmed} & \textbf{Actor} & \textbf{Chameleon} & \textbf{Squirrel} & \textbf{Cornell} & \textbf{Texas} & \textbf{Wisconsin} \\ 
\Xhline{2\arrayrulewidth}
                        & \ Nodes  & 2,708  & 3,327   & 19,717 & 7,600 & 2,277  & 5,201 & 183 & 183 & 251 \\
                        & \ Edges         & 10,558  & 9,104  & 88,648   & 25,944 & 33,824  & 211,872 & 295 & 309 & 499 \\
                        & \ Features       & 1,433  & 3,703  & 500   & 931 & 2,325  & 2,089 & 1,703 & 1,703 & 1,703 \\
                        & \ Classes        & 7  & 6  & 3     & 5  & 5  & 5 & 5 & 5 & 5 \\
\Xhline{2\arrayrulewidth}
\end{tabular}
\end{adjustbox}
\end{table*}

\section{Details of Datasets and Baselines} \label{det_dat_bas}
\paragraph{Datasets.}
To comprehensively evaluate model performance under varying structural regimes, we conduct experiments on nine widely adopted benchmark datasets, covering both homophilic and heterophilic graphs that are described in Table \ref{dataset}. The datasets include three classical citation networks: Cora, Citeseer, and Pubmed \cite{kipf2017semi}, characterized by high homophily ($\mathcal{G}_h \in [0.74, 0.81]$), and six additional benchmarks that exhibit strong heterophily: Actor, Chameleon, Squirrel, Cornell, Texas, and Wisconsin \cite{rozemberczki2019gemsec}. Actor is a co-occurrence network derived from film actor collaborations, while Chameleon and Squirrel are Wikipedia page networks with low homophily and rich structural diversity. Cornell, Texas, and Wisconsin are WebKB networks with extremely low $\mathcal{G}_h$ (0.06-0.16), often used as canonical heterophily benchmarks. We follow the same data splits and preprocessing protocols as prior heterophily literature \cite{zhu2020beyond, bo2021beyond, li2022finding}, including fixed 20 labeled nodes per class for training and the remaining nodes split evenly into validation and test sets.

\paragraph{Baselines.}
To ensure fair and comprehensive comparison, we benchmark our method against 15 representative GNN baselines that cover classical, heterophily-aware, spectral-filtering, and recent state-of-the-art methods.  

\begin{itemize}
    \item \textbf{Classical GNNs:}
    GCN \cite{kipf2017semi} and GAT \cite{velivckovic2018graph} are the standard message-passing architectures, serving as strong baselines in homophilic settings.
    \item \textbf{Heterophily-aware propagation:}
    H\textsubscript{2}GCN \cite{zhu2020beyond} separates ego and neighbor aggregation to alleviate bias, 
    GPRGNN \cite{chien2021adaptive} learns personalized propagation weights, and 
    FAGCN \cite{bo2021beyond} adaptively mixes low- and high-frequency signals.
    \item \textbf{Spectral and signed filtering:}
    SIGN \cite{frasca2020sign} precomputes propagation features, 
    MagNet \cite{zhang2021magnet} introduces magnetic Laplacians for directional structure, GCNII \cite{chen2020simple} incorporates identity mapping to counter over-smoothing, and L2DGCN \cite{dingl2dgcn} mitigates bias-degree via learnable graph augmentation. 
    \item \textbf{Adaptive and structure-enhanced models:}
    ACM-GCN \cite{luan2022revisiting} uses channel mixing, 
    GloGNN \cite{li2022finding} adds global nodes to enhance long-range propagation,
    Auto-HeG \cite{zheng2023auto} automates heterophily architecture search,
    DirGNN \cite{rossi2024edge} explicitly models directed edges,
    PCNet \cite{li2024pc} filters homophilic signals in heterophilic graphs,
    TFE-GNN \cite{duan2024unifying} integrates topology and feature filtering,
    and CGNN \cite{pmlr-v267-zhuo25a} uses contrastive objectives to improve generalization.
\end{itemize}

This diverse set of baselines allows us to systematically compare GESC with (i) early message-passing GNNs, (ii) heterophily-specialized models, and (iii) recent spectral and structural advances, evaluating its effectiveness across a wide spectrum of graph conditions.

\end{document}